%% file: main.tex
\PassOptionsToPackage{table}{xcolor}
\documentclass[10pt,twocolumn,letterpaper]{article}

\usepackage{cvpr}              

\usepackage{microtype}
\usepackage{graphicx}
\usepackage{subcaption}
\usepackage{booktabs}
\usepackage{amsmath}
\usepackage{fvextra}
\usepackage{amssymb}
\usepackage{mathtools}
\usepackage{amsthm}
\usepackage{listings}
\usepackage{longtable}
\usepackage{multirow}
\usepackage{colortbl}
\usepackage{pgf}
\usepackage{enumitem}
\usepackage{tcolorbox}
\tcbuselibrary{skins, breakable}
\usepackage{array}
\usepackage{makecell}
\usepackage{tabularx}
\usepackage{pifont}
\usepackage{textcomp}
\usepackage{cuted}
\usepackage[disable,textsize=tiny]{todonotes}

\definecolor{cvprblue}{rgb}{0.21,0.49,0.74}
\usepackage[pagebackref,breaklinks,colorlinks,allcolors=cvprblue]{hyperref}
\usepackage[capitalize,noabbrev]{cleveref}

\def\paperID{*****}
\def\confName{CVPR}
\def\confYear{2026}

\title{MMHBench: A Multi-Perspective Benchmark for Mental Health Understanding in Long-Form Videos}

\author{
Jinpeng Hu$^{1}$ \quad
Erqiang Wang$^{1}$ \quad
Shan Wang$^{2}$ \\
Zhuo Li$^{3}$ \quad
Peipei Song$^{4}$ \quad
Xun Yang$^{4}$\quad
Meng Wang$^{1}$\\
$^{1}$Hefei University of Technology \quad
$^{2}$Faculty of Psychology, Beijing Normal University \\
$^{3}$The Chinese University of Hong Kong, Shenzhen \\
$^{4}$University of Science and Technology of China \\
{\tt\small
jinpenghu@hfut.edu.cn,
erqiangwang@mail.hfut.edu.cn}\\
}

\theoremstyle{plain}

\theoremstyle{definition}

\theoremstyle{remark}

\newcommand{\cmark}{\textcolor{green!60!black}{\ding{51}}}
\newcommand{\xmark}{\textcolor{red!70!black}{\ding{55}}}

\newcolumntype{P}[1]{>{\centering\arraybackslash}m{#1}}
\newcolumntype{Z}{>{\centering\arraybackslash}X}

\newif\ifcomments
\commentstrue

\definecolor{highcolor}{HTML}{137333}
\definecolor{lowcolor}{HTML}{FFFFFF}

\definecolor{primaryparticipanttextcolor}{RGB}{57,94,95}
\definecolor{psychologisttextcolor}{RGB}{115,128,125}
\definecolor{familymembertextcolor}{RGB}{178,157,196}
\definecolor{othertextcolor}{RGB}{189,148,135}
\definecolor{8B}{RGB}{184,224,212}
\definecolor{14B}{RGB}{149,184,209}
\definecolor{2B}{RGB}{229,212,239}
\definecolor{other}{RGB}{255,214,165}
\definecolor{mental}{RGB}{229,212,239}

\newcommand{\benchmarkname}{MMHBench}
\newcommand{\totalVideoNumber}{268}
\newcommand{\totalCandidateVideoNumber}{1,600}
\newcommand{\totalQuestionNumber}{2,184}
\newcommand{\totalThirdQuestionNum}{605}
\newcommand{\totalFirstQuestionNum}{1,579}

\newcommand{\Rone}{LES}
\newcommand{\Rtwo}{MHP}
\newcommand{\Rthree}{FM}
\newcommand{\Rfour}{Others}

\newcommand{\MLLMNum}{22}

\newcommand{\Ronespecific}{Lived Experience Subject}
\newcommand{\Rtwospecific}{Mental Health Professional}
\newcommand{\Rthreespecific}{Family Member}
\newcommand{\Rfourspecific}{Others} 

\begin{document}
\maketitle
\begin{strip}
\centering
\includegraphics[width=0.99\textwidth, trim=0 0 0 0]{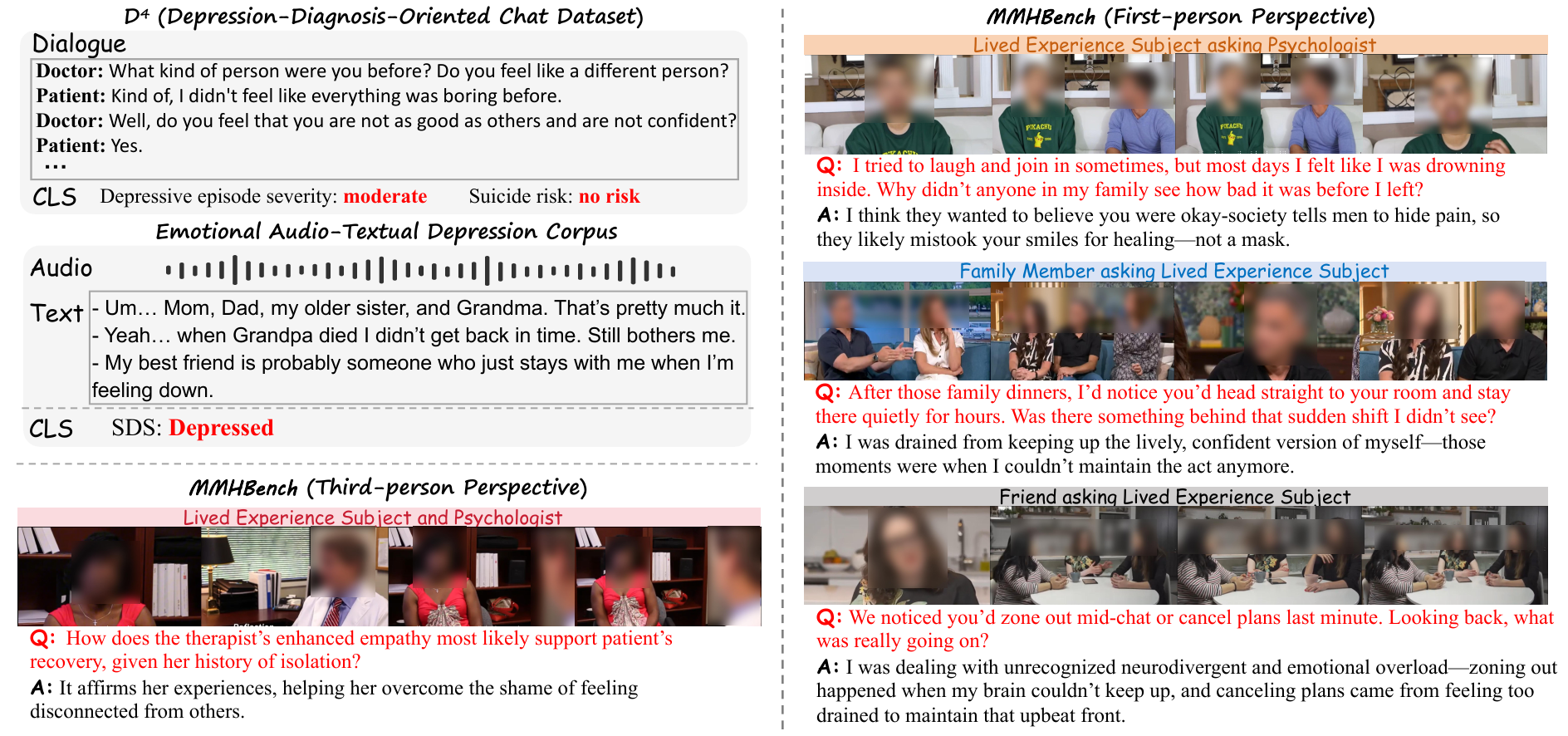}
\captionof{figure}{Examples from {\benchmarkname}, including both first-person and third-person questions.}
\label{fig:intro}
\end{strip}

\begin{abstract}

Mental health understanding in long-form videos requires nuanced reasoning over observable behavior, interpersonal context, and latent psychological states. Existing benchmarks largely reduce this task to coarse-grained classification, providing limited insight into whether models truly understand psychological phenomena or rely on superficial correlations. To address this limitation, we introduce MMHBench, a comprehensive multimodal benchmark for multi-perspective mental health understanding, comprising 268 long-form videos and 2,184 carefully curated questions. MMHBench organizes the evaluation into two complementary settings: (1) third-person assessment, consisting of 605 questions that focus on the interpretation of observable behaviors and multimodal evidence, and (2) first-person perspective-taking, comprising 1,579 questions that require perspective-conditioned reasoning to identify the interpretation of the mental state supported by the available multimodal evidence. We propose a Multi-Agent Question Generation (MAQG) framework that simulates diverse social roles to synthesize questions from multiple perspectives. The generated questions are refined through multi-role feedback and iterative optimization, followed by expert-guided verification to ensure quality and validity. Extensive evaluation of 22 representative multimodal large language models (MLLMs), spanning both open-source and leading closed-source models, demonstrates that long-form video mental health understanding remains highly challenging.

\end{abstract}

\section{Introduction}

Mental health has become a critical global concern, creating a growing demand for timely and accessible psychological support.
However, shortages of qualified professionals and persistently low help-seeking rates continue to limit access to effective interventions.
Machine learning has consequently been explored for mental health screening and analysis across social media, clinical interviews, and multimodal behavioral data~\cite{yang2017multimodal,ansari2022ensemble,orabi2018deep,islam2018depression,lin2020sensemood,hu2026agentmental}.
Early studies primarily employed conventional neural architectures for depression detection~\cite{orabi2018deep}, while recent approaches increasingly incorporate large language models (LLMs) to extract clinically relevant information and support more interpretable mental health analysis~\cite{lan2025depression,chen2024depression,openai2024gpt4technicalreport,comanici2025gemini,hu2026agentmental}.

Parallel to these method advancements, the curation of standardized datasets has been instrumental in benchmarking progress \cite{gratch-etal-2014-distress, shen2022automatic,yao2022d4,zou2022semi, hyun2024smile, fu2025first, he2025lmvd}.
Representative resources such as DAIC-WOZ~\cite{gratch-etal-2014-distress}, EATD-Corpus~\cite{shen2022automatic}, and MPDD~\cite{fu2025first} provide multimodal recordings or derived features for mental health screening, disorder detection, and severity estimation.
Despite their importance, existing benchmarks predominantly rely on coarse-grained classification paradigms, reducing complex psychological profiles to binary or ordinal labels (e.g., ``depressed'' vs. ``non-depressed"), as illustrated in Figure~\ref{fig:intro}.
Such formulations provide limited insight into whether a model can interpret the emotional, behavioral, and interpersonal evidence underlying its predictions.
Real-world mental health understanding, however, requires richer reasoning across different perspectives.
Most existing benchmarks fail to distinguish between first-person subjective experiences and third-person observations.
Third-person reasoning primarily requires an external interpretation of observable evidence, whereas first-person perspective-taking requires models to reason from the subjective standpoint of a specific individual and understand implicit experiences, emotions, and intentions.
Moreover, many widely used mental health benchmarks release only pre-extracted multimodal features, such as OpenFace or OpenSmile representations, rather than original video content, limiting the ability to evaluate multimodal perception and reasoning in multimodal large language models (MLLMs).

\input{table/existing_dataset}

To address these limitations and move beyond surface-level classification, we introduce a novel \textbf{M}ultimodal \textbf{M}ental \textbf{H}ealth understanding benchmark ({\benchmarkname}), comprising {\totalVideoNumber} long-form videos and {\totalQuestionNumber} carefully curated questions.
Unlike preceding benchmarks that prioritize diagnostic labels, {\benchmarkname} is designed to assess fine-grained, perspective-aware understanding in long videos through meticulously curated video-grounded questions.
The benchmark contains two complementary question settings.
Third-person questions assess whether models can identify and integrate observable behavioral, contextual, and interactional evidence from an external viewpoint.
Conversely, first-person questions require models to reason from the perspective of a specific individual in the video, thereby evaluating their ability to identify plausible subjective experiences and psychological states grounded in multimodal evidence.
Together, these settings enable a systematic analysis of the distinction between external observation and perspective-conditioned psychological reasoning.
A more comprehensive comparison between {\benchmarkname} and existing benchmarks is provided in Table~\ref{tab:benchmark_comparisons}.
To support scalable yet reliable benchmark construction, we propose a Multi-Agent Question Generation (MAQG) framework.
We first employ a human-in-the-loop iterative process to refine the generation prompts and expand the seed question pool.
Then, we utilize a perception agent to extract granular contextual cues and character profiles to initialize a set of role-conditioned agents.
These role-conditioned agents simulate diverse social roles to generate questions and provide complementary peer feedback.
Subsequently, an iterative refinement stage integrates textual and multimodal evaluation agents to assess question validity and multimodal grounding before a revision agent consolidates feedback to update the content accordingly.
The resulting candidates are further processed through answer leakage filtering and expert-guided verification to ensure data validity.
We extensively evaluate {\MLLMNum} representative MLLMs, spanning both open-source and leading closed-source models.
The results demonstrate that \textit{long-video mental health understanding remains a \textbf{substantial challenge} for current MLLMs}.
In particular, most models achieve substantially lower accuracy on first-person questions than on third-person questions, revealing pronounced limitations in evidence-grounded psychological reasoning from specific role perspectives.

\section{Related Work}
Many benchmarks have been proposed to assess MLLMs' long-context capabilities for long video understanding.
For instance, some recent benchmarks \citep{fu2025video, wang2025lvbench,wu2024longvideobench,ataallah2025infinibench} aim at general-purpose evaluation by assessing MLLMs’ long-context capabilities across multiple aspects, including fine-grained perception, reasoning, and holistic understanding.
Furthermore, MLVU \citep{zhou2025mlvu} and the work of \citet{song2025video} focus on multi-disciplinary, knowledge-intensive evaluation for long video understanding, offering broader subject coverage for assessing MLLMs in long-context settings.
%
Related research on multimodal affective computing further examines models' ability to recognize emotions by integrating multimodal cues~\citep{zadeh2018multimodal,poria2019meld,zhao2022m3ed,hu2025beyond,dai-etal-2026-tears,song2024emotional}.
However, mental health understanding extends beyond affect recognition, requiring models to reason about more complex psychological states, behavioral patterns, and interpersonal contexts.
Broadly, existing approaches can be categorized into text-centric evaluation
\citep{zhai2025mentalglm,qiu2025deepwell,ma2025psyscam,sathvik2025m,zhang2025auradial} and multimodal analysis \citep{yoon2022d,haydarov2025towards}.
DeepWell-Adol \cite{qiu2025deepwell} and AURADIAL \cite{zhang2025auradial} introduce large-scale Chinese counseling dialogue datasets designed to support mental health assessment and intervention in multi-turn consultation settings, emphasizing empathetic and human-like responses grounded in real-world user queries.
To capture non-verbal signals, D-Vlog \cite{yoon2022d} analyzes in-the-wild vlog videos for depression detection, while LMVD \cite{he2025lmvd} targets depression detection in the wild and publicly releases pre-extracted multimodal cues to facilitate feature-based classification.
However, these studies often remain surface-level and fail to enable in-depth, multi-perspective evaluation of psychological states.
To bridge this gap, we propose {\benchmarkname}, a multi-perspective benchmark that explicitly distinguishes and jointly evaluates first-person subjective experience and third-person objective observation.

\begin{figure*}[t!]
    \centering
    \includegraphics[width=1\linewidth, trim=0 0 0 10]{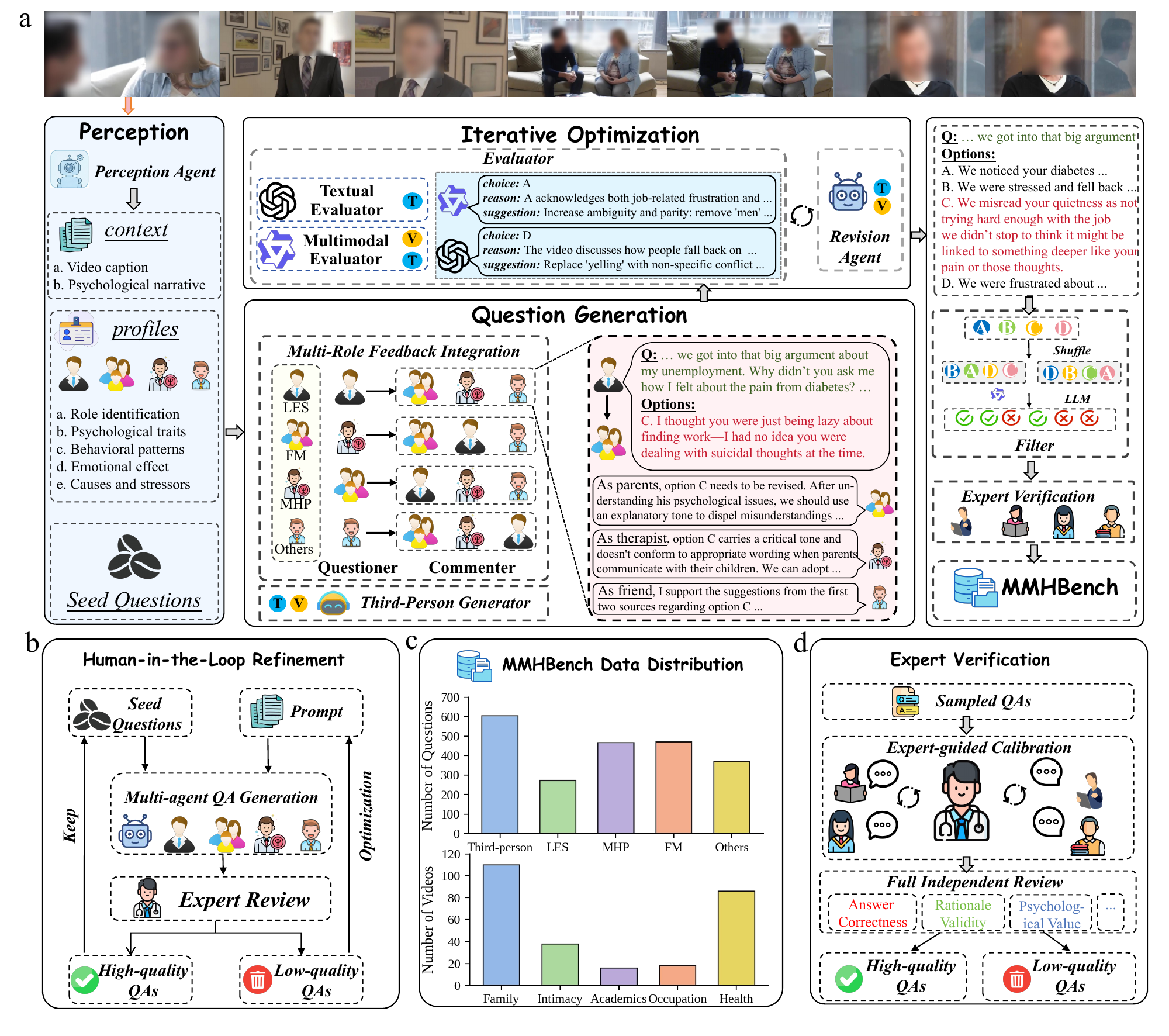}
    \vspace{-1em}
    \caption{
        An overview of MMHBench.
        \textbf{a.} The data construction process employs a multi-agent framework, including perception, multi-role feedback integration, iterative optimization, visual grounding filtering, and manual verification.
        {b.} Human-in-the-loop prompt refinement and seed pool expansion.
        {c.} Distributions of video types and perspective categories.
        {d.} Expert verification.
        }
    \label{fig:framework}
    \vspace{-1.5em}
\end{figure*}

\section{{\benchmarkname}}
In this section, we introduce the {\benchmarkname} benchmark and present its overall framework shown in Figure~\ref{fig:framework}, covering data collection, question generation (Figure~\ref{fig:framework}(a) and (b)), data statistics (Figure~\ref{fig:framework}(c)), and expert verification (Figure~\ref{fig:framework}(d)).

\subsection{Data Collection}
%
Our dataset is collected from publicly available psychological interview and self-narrated videos from multiple online video-sharing platforms, including YouTube, Bilibili, and other public video-sharing platforms.
These videos contain discussions of emotional states, personal experiences, interpersonal relations, coping behaviors, and psychological difficulties, providing rich multimodal evidence for mental health related video understanding.
We initially collect approximately {\totalCandidateVideoNumber} candidate videos to ensure diversity in content, recording conditions, speaker backgrounds, and interaction styles.
To ensure the quality and suitability of the collected videos, we employ a structured human review process involving five reviewers under the supervision of a licensed psychological counselor.
Each candidate video is assessed according to five dimensions: Topic Relevance, Psychological Process Evidence, Narrative Completeness, Video Duration, and Ethical Appropriateness.
The detailed rubric is provided in Appendix~\ref{app:video_filtering_rubric}.
Prior to formal annotation, all reviewers complete a calibration phase on a pilot set of videos to establish a shared understanding of the evaluation guidelines.
Subsequently, each candidate video is independently assessed by a reviewer.
Following content-level filtering, we further remove non-informative segments that do not contribute to substantive mental health-related content, including opening and closing credits, platform-specific introductions, advertisements, and unrelated entertainment segments.
Through this multi-stage screening and cleaning procedure, we obtain {\totalVideoNumber} high-quality videos that constitute the final benchmark.
These videos span five thematic domains: Family, Intimacy, Academics, Occupation, and Health, as shown in Figure~\ref{fig:framework}(c).

\subsection{Question Types}
We stratify our question design along two narrative perspectives: first-person and third-person.

\noindent \textbf{Third-Person Perspective.} Third-person inquiries emphasize an exocentric interpretation of video content grounded in explicit multimodal observations.
These questions require the model to synthesize observable external cues to achieve objective scene comprehension, such as behavior and situational context.
By functioning as an external observer, the model maps perceived physical behaviors to coherent semantic interpretations.

\noindent \textbf{First-Person Perspective.} 
Conversely, first-person questions adopt an egocentric cognitive modeling paradigm that is fundamental to mental health reasoning, requiring the model to reason from the perspective of a specific individual involved in the video.
Moving beyond the purely observational nature of third-person tasks, this category emphasizes reasoning over latent internal states that are not directly manifested in behavior.
Based on the conversational dynamics within these videos, we classify first-person questions into four distinct roles, {\Ronespecific} ({\Rone}), {\Rtwospecific} ({\Rtwo}), {\Rthreespecific} ({\Rthree}), and {\Rfourspecific}.
The detailed descriptions of these roles are shown in Appendix \ref{details_of_roles}.
%
%
\subsection{Human-in-the-Loop Refinement}
\label{seed_question_construction}

We first construct an initial seed pool consisting of three complementary types of questions.
\begin{itemize}[leftmargin=*]
\setlength{\topsep}{0pt}
\setlength{\itemsep}{0pt}
\setlength{\parsep}{0pt}
\setlength{\parskip}{0pt}
    \item We extract naturalistic questions from publicly available, high-fidelity videos of real-world interactions among individuals with lived experience, clinicians, family members, and others.
    \item We recruit two licensed psychological counselors to author a set of domain-relevant questions.
    \item We manually craft perception-focused questions targeting fine-grained visual and behavioral cues relevant to mental health.
\end{itemize}

The initial seed pool contains 105 questions in total. To optimize the generation prompts and expand the seed pool, we adopt a human-in-the-loop generation and refinement process, as illustrated in Figure~\ref{fig:framework}(b). 
At each iteration, we randomly sample 3 examples from the current seed pool as in-context demonstrations for MAQG. 
These examples provide structural guidance for question generation, distractor construction, gold answer calibration, and question analysis.
Conditioned on the sampled demonstrations and the current generation prompts, MAQG employs multiple role-conditioned agents to generate a new batch of 12 candidate question-option pairs. 
Each generated candidate question-option pair is assigned to two licensed psychological counselors.
After completing independent evaluations, the two experts discuss any issues or discrepancies identified and jointly decide whether the candidate questions should be retained, modified, or discarded.
Candidates that meet the quality standards are retained and added to the seed pool, thereby progressively enhancing the pool's breadth and diversity.
Following review by experts, candidate content with correctable issues may be added to the seed pool after revision, whereas content with unresolvable quality issues is discarded.
The reviewers analyze rejected and modified candidates to identify prevalent generation failures, such as insufficient multimodal grounding of the answers and rationales.
Based on these recurring failure patterns, we refine the generation prompts through targeted revisions, including role instructions, evidence-grounding constraints, and guidance for question formulation and distractor construction.
The revised prompts are then used in the next iteration to generate a new batch of candidates.
This dual-feedback mechanism allows accepted candidates to expand the seed pool, while modified and rejected candidates directly drive prompt refinement.
The expanded seed examples are subsequently subjected to the answer leakage filter and the expert verification procedure described in Section~\ref{sec:human_verify}. 
Finally, the iterative process yields an expanded seed pool containing 215 high-quality examples and a refined set of generation prompts, which respectively serve as in-context demonstrations and generation
instructions for downstream benchmark construction. 

\subsection{Multi-Agent Data Generation Framework}
To generate the aforementioned questions in a structured and controllable manner, we design a Multi-Agent Question Generation (MAQG) framework that decomposes the generation process into several specialized agents with complementary functions, reducing the substantial human effort required for question generation.
This modular design supports specialized processing and staged refinement.
Specifically, our data synthesis framework comprises a perception agent $\text{Agent}_\text{perc}$, multiple role-playing agents that emulate the perspectives of {{\Rone}}, {{\Rtwo}}, {{\Rthree}}, and {{\Rfour}}, as well as a dedicated third-person question generation agent.
To ensure the technical integrity and cognitive grounding of the synthetic tasks, we also incorporate a \textbf{t}extual \textbf{e}valuator $\text{Agent}_{\text{te}}$, a \textbf{m}ultimodal \textbf{e}valuator $\text{Agent}_{\text{me}}$ and a \textbf{re}vision agent $\text{Agent}_{\text{re}}$ that jointly provide automated quality control and refinement.

\noindent \textbf{Multimodal Perception.}
Our framework begins with the perception agent $\text{Agent}_\text{perc}$, which serves as the foundational observer. $\text{Agent}_\text{perc}$ processes the raw video $\mathcal{V}$ to extract salient spatiotemporal context and identify the social roles within the $\mathcal{V}$ instructed by $p_\text{perc}$ formalized as:
\begin{equation}
    \{C, R\} = \text{Agent}_\text{perc}(p_\text{perc}, \mathcal{V}),
    \label{eq:perc}
\end{equation}
where $C$ denotes the extracted video information, including salient events, emotional expressions, video captions, and an overall video summary of $\mathcal{V}$.
$R=\{r_{{{\Rone}}}, r_{{{\Rtwo}}}, r_{{{\Rthree}}}, r_{{{\Rfour}}}\}$ represents available role profiles associated with $\mathcal{V}$, corresponding to {{\Rone}}, {{\Rtwo}}, {{\Rthree}} and {{\Rfour}} roles, respectively.
Not all videos contain all four roles, with some videos including only a subset of these role types.
This shared representation $C$ serves as the common context for all subsequent agents, ensuring consistency across generated questions.

\noindent \textbf{Question Generation.}
We use the refined seed questions described in Section~\ref{seed_question_construction} as in-context demonstrations to guide automatic question generation.
Subsequently, role-playing agents are dynamically instantiated based on the available role profiles $R$, the video $\mathcal{V}$, and the distilled context $C$.
Specifically, each role-playing agent is explicitly prompted to simulate the distinct perspective, communicative intent, and cognitive reasoning style inherent to its designated role.
This configuration directs the agents to synthesize inquiries that align with the specific epistemic concerns of each character type.
These role-playing agents are further used to generate diverse types of first-person questions along with their corresponding options.
For example, a question from the perspective of the {\Ronespecific} is formalized as:
\begin{equation}
    \{q, O\} = \text{Agent}_{\text{LES}}(p_{\text{LES}}, r_{\text{LES}}, C, \mathcal{V}),
\end{equation}
where $p_{{\Rone}}$ denotes the prompt used to guide the LLM to simulate {\Ronespecific}, $r_{\text{LES}}$ is the role profile of LES, $q$ is the generated question and $O=\{o_{1},o_{2},o_{3},\cdots\}$ represents the set of distractors and the provisional reference answer, which will be subsequently validated through multiple verification stages.
In addition, for third-person questions that do not require explicit role simulation, we employ a separate question generation agent, which also leverages seed questions as guidance to directly generate questions grounded in the video $\mathcal{V}$ and the distilled contextual representation $C$.

\noindent \textbf{Multi-Role Feedback Integration.}
To ensure epistemic consistency and social authenticity within the generated first-person questions, we implement a multi-role collaborative refinement protocol.
Following the initial generation of a question–option pair ${Q, O}$ by a primary role-playing agent, the remaining role-playing agents function as auxiliary critics, systematically optimizing the instance by providing feedback grounded in their unique epistemic viewpoint.
%
%
They examine the social plausibility of distractor options, the relational appropriateness of the ground-truth answer, and the consistency between the question and the underlying multimodal evidence.
The interaction among role-playing agents produces a collection of role-specific feedback signals, formalized as:
\begin{equation}
    \mathcal{F} = \{f_{r_{i}} \mid r_{i} \in R \setminus \{r_{\text{init}}\} \},
\end{equation}
where $r_{\text{init}}$ denotes the role responsible for the initial pair generation and $f_{r_{i}}$ is the feedback provided by the agent associated with role $r_i$.
To ensure balanced contribution across perspectives, role-playing agents alternate between serving as the primary question generator and as feedback providers. 
In each generation round, one role-playing agent produces question–option pairs, while the remaining agents critique these pairs from their perspective.

Then, the revision agent $\text{Agent}_{\text{re}}$ incorporates $\mathcal{F}$ as additional input and revises both the question and its corresponding options, producing higher-quality, role-consistent pairs:
\begin{equation}
    \{q, O\} = \text{Agent}_{\text{re}}(p_{\text{re}}, \mathcal{F}, q, O),
\end{equation}
where $p_{\text{re}}$ is the prompt guiding the revision agent.

\noindent \textbf{Iterative Optimization.}
To reduce triviality and linguistic leakage, we employ a textual evaluator agent $\text{Agent}_{\text{te}}$ and a multimodal evaluator agent $\text{Agent}_{\text{me}}$.
The former evaluates questions using text alone, while the latter considers both textual and visual information to ensure that the questions are challenging and grounded in the source video.
Specifically, for a given pair $\{q, O\}$, $\text{Agent}_{\text{te}}$ first attempts to predict the answer in a blind setting, without access to the video.
If the predicted answer is correct, this outcome suggests that the question can be resolved solely from textual cues and is therefore insufficiently grounded in multimodal evidence.
In this case, $\text{Agent}_{\text{te}}$ provides targeted suggestions $s_{\text{te}}$ on how the question can be revised to require stronger multimodal grounding, encouraging closer alignment with the underlying video evidence.
In contrast, the multimodal evaluator $\text{Agent}_{\text{me}}$ evaluates the pair $\{q, O\}$ in conjunction with the source video. 
Beyond validating the accuracy of the ground truth, its primary objectives are to generate specialized suggestions $s_{\text{me}}$ aimed at increasing the difficulty of the questions and the plausibility of the distractor options.
The suggestions $s_{\text{te}}$ and $s_{\text{me}}$ are jointly provided to the revision agent $\text{Agent}_{\text{re}}$ as $s=[s_{\text{te}}, s_{\text{me}}]$, which updates the question and its options accordingly.
To ensure the quality of generated data, the framework implements an iterative optimization loop.
Following each revision, the updated question is then re-evaluated by $\text{Agent}_{\text{te}}$ and $\text{Agent}_{\text{me}}$.
Their subsequent feedback is iteratively incorporated by $\text{Agent}_{\text{re}}$ to incrementally enhance the complexity and grounding. 
\begin{equation}
    \{q^{(t+1)}, {O}^{(t+1)}\} = \text{Agent}_{\text{re}}(p_{rev}, s^{(t)}, q^{(t)}, {O}^{(t)}),
    \label{iterative_optimization}
\end{equation}
where $t$ denotes the iteration index, and $p_{\text{rev}}$ denotes the revision prompt used to guide the iterative optimization process.
This optimization loop continues for a predefined maximum number of iterations, or the updated pair satisfies the evaluation imposed by the evaluator and revision agents.

\input{table/overall_performance}

\subsection{Leakage Filtering and Expert Verification}
\label{sec:human_verify}
A reliable multimodal benchmark should prevent questions from being consistently answered using cues contained only in the question and answer options.
Although the textual evaluator agent provides an initial check against unimodal shortcuts during generation, residual answer leakage may still arise from overly explicit linguistic cues, implausible distractors, or option patterns that make the correct answer identifiable without access to the video.
To mitigate this issue, we apply a post-generation answer leakage filter.
For each question-option pair, two different strong text-only LLMs each perform three independent trials with randomized answer-option orders, receiving only the question and shuffled options without access to the corresponding video, resulting in six trials in total.
If the ground-truth answer is correctly selected in at least four of the six trials, the item is considered potentially affected by answer leakage or language-only shortcuts and is removed.
This procedure reduces answer-position bias and identifies questions that remain stably solvable without observing the video content.
To further ensure data quality, we recruit two additional licensed psychological counselors, resulting in a team of four experts for verification.
The two experts involved in prompt optimization and seed pool expansion in Section~\ref{seed_question_construction} first develop a five-dimensional rubric based on their prior review experience.
Before formal annotation, all four experts review the question format, evaluation dimensions, and a set of warm-up examples to establish a shared understanding of the rubric.
We then randomly sample 50 generated question-option pairs as the first calibration subset, which all four experts annotate independently.
After the first round, the experts review items with substantial rating discrepancies, discuss the sources of disagreement, and refine the rubric accordingly.
Detailed definitions and scoring criteria for these dimensions are provided in Appendix~\ref{expert_verification_rubric}.
To assess whether the refined rubric can be consistently applied to unseen items, we randomly sample a second disjoint subset of 200 question-option pairs.
All four experts independently annotate this second calibration subset using the refined rubric.
We compute Krippendorff's $\alpha$ for these evaluation dimensions in the second calibration round to quantify inter-rater agreement among the four counselors, with results shown in Appendix~\ref{app:human_agreement}.
After calibration, we partition all remaining question-option pairs into four disjoint subsets and assign one subset to each expert.
Each expert evaluates the assigned subset using the same rubric and we retain question-option pairs that satisfy the rubric. 
Each revised item is subsequently reviewed by another expert and retained only if it meets the established criteria.

\subsection{Data Statistics}

In this section, we present the statistical overview of {\benchmarkname}. 
Figure~\ref{fig:framework}(c) presents the data statistics of {\benchmarkname}.
In terms of scale, {\benchmarkname} comprises {\totalVideoNumber} videos spanning five primary thematic domains, paired with {\totalQuestionNumber} multiple-choice questions, including {\totalThirdQuestionNum} third-person questions and {\totalFirstQuestionNum} first-person questions.
These videos are categorized into five primary thematic domains, namely Family, Intimacy, Academics, Occupation, and Health, with their distribution shown in Figure~\ref{fig:framework}(c).
Within the first-person subset, questions raised by {{\Rone}}, {{\Rtwo}}, {{\Rthree}}, and {{\Rfour}} account for 272, 467, 470, and 370 instances, respectively.
This distribution supports fine-grained evaluation of multimodal mental health reasoning from diverse interpersonal perspectives.
This composition enables the benchmark to evaluate both observer-perspective reasoning and role-conditioned mental state inference.

\section{Experiments}

\subsection{Baselines and Evaluation Metrics}
We evaluate various representative MLLMs across a wide range of scales using accuracy (\%) as the primary metric. The evaluated open-source MLLMs include Qwen2.5-VL series \cite{bai2025qwen2}, 
Qwen3-VL series~\cite{bai2025qwen3vltechnicalreport}, 
Qwen3.5  \cite{qwen3.5}, 
Qwen3.6-35B-A3B \cite{qwen36_35b_a3b}, 
InternVL3.5 series \cite{wang2025internvl3}, 
Ministral-3 series \cite{liu2026ministral}, 
Gemma-4 series,
MiniCPM-V-4 \cite{yao2024minicpm}, 
MiniCPM-V-4\_5 \cite{yu2026minicpm}, 
and MiniCPM-V-2.6 \cite{yao2024minicpm}.
The closed-source MLLMs comprise GPT-5.5,
Qwen3.6-Plus \cite{qwen36plus} and Qwen3.7-Plus \cite{qwen37}.
Implementation details are shown in Appendix \ref{Implementation_Details}.
%

\begin{figure*}[t]
\centering
 \includegraphics[width=1\linewidth, trim=0 0 0 10]{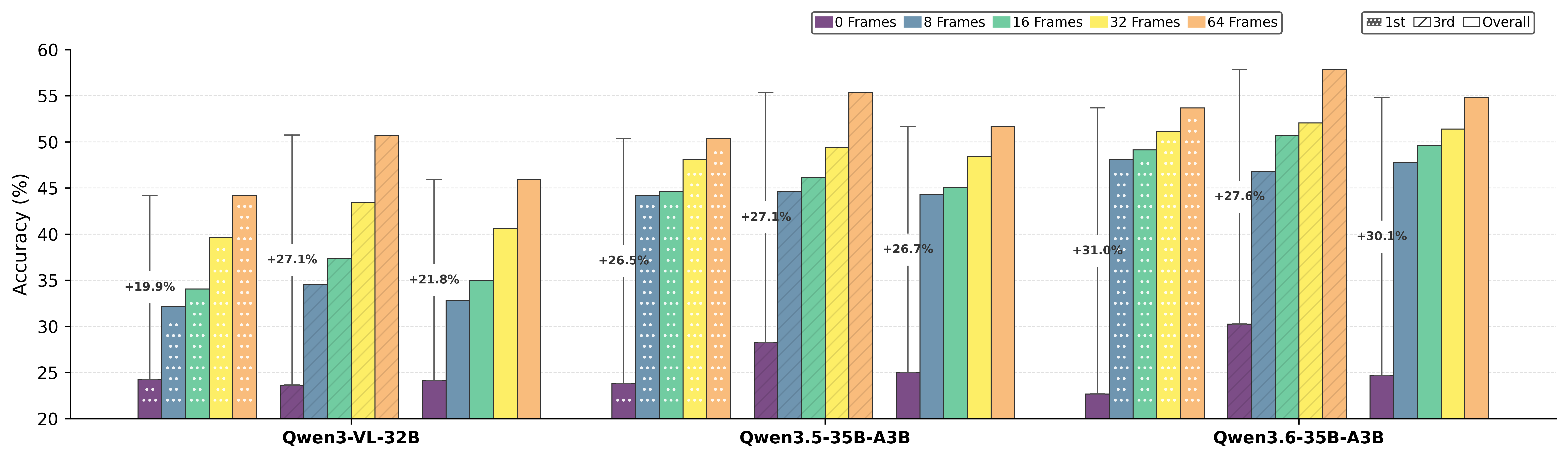}
\vskip -0.5em
\caption{Impact of input frame density on model performance.}
\label{fig:video-frame}
\vspace{-1em} 
\end{figure*}

%

\subsection{Main Results}
To explore the performance of different models on {\benchmarkname}, we report the overall accuracy (\%) in Table~\ref{tab:overall_results} and the performance breakdown across different question types.

\noindent \textbf{Overall Performance Analysis.}
Among these models, \textsc{GPT-5.5} achieves the strongest overall performance, indicating that large-scale proprietary pre-training employed in closed-source development significantly enhances a model’s ability to navigate the nuances of human social behavior.
Within the same model family, larger variants usually outperform their smaller counterparts, which can be attributed to their stronger capacity for long-context reasoning and more effective integration of multimodal information over extended temporal horizons.
Beyond intra-family scaling effects, Qwen3.5 also demonstrates competitive performance compared with other model families at similar parameter scales.
This overall performance margin underscores the effectiveness of Qwen3.5 in capturing and integrating complex multimodal cues critical for mental health understanding.
%
\input{table/reasoning_quality}
\noindent \textbf{Impact of Inferential Perspective.} 
Comparative results between first-person and third-person questions show that first-person questions are generally more challenging, yielding lower accuracy across most evaluated models.
This performance gap arises because third-person questions primarily assess comprehension of observable events and behaviors, whereas first-person questions require deeper introspective reasoning to infer latent mental states, emotions, and subjective intentions.
These results indicate that mental health understanding from a first-person perspective remains difficult for current models, highlighting the importance of explicitly evaluating perspective-aware mental health reasoning.

\noindent \textbf{Role-Dependent Performance Analysis.}
To analyze how social roles influence model performance, we evaluate performance along two dimensions: the role of the questioner and the role assumed by the responder.
When considering the role of the questioner, we find that questions posed from the {\Ronespecific} perspective are generally more challenging than those from other roles. 
These questions require the model to respond as an interlocutor addressing an individual experiencing psychological distress, thereby demanding empathetic understanding and reasoning over implicit mental states rather than surface-level content.
A similar difficulty emerges when conditioning responses on the {\Rthreespecific} perspective, which yields lower accuracy across most evaluated models.
This setting requires models to reason from indirect behavioral evidence and relational context, which increases the ambiguity of interpersonal and psychological inference.

\noindent \textbf{Reasoning Quality Evaluation.}
Beyond answer accuracy, we evaluate model-generated explanations along four dimensions: key event (KE), core psychological mechanism (CPM), option explanation (OE), and hallucination.
Hallucination assesses whether explanations avoid unsupported diagnoses, incorrect causal attributions, and claims not grounded in the video.
GPT-5.5 serves as the automatic judge, with the detailed scoring rubric provided in Appendix~\ref{app:reasoning_quality_rubric}.
To validate the automatic evaluation, we collect ratings from a licensed psychological counselor for 130 explanations randomly sampled from those generated by Qwen3.6-35B-A3B.
We report Spearman's $\rho$ to assess agreement between GPT-5.5 judgments and the expert ratings, as detailed in Appendix~\ref{app:gpt_human_agreement}.
As shown in Table~\ref{tab:reasoning_quality}, core mechanism identification and option explanation quality exhibit closely aligned model rankings, suggesting that coherent option-level reasoning is associated with accurate identification of the underlying psychological mechanism.
Qwen3.6-35B-A3B achieves the best performance across most dimensions, consistent with its leading overall accuracy in Table~\ref{tab:overall_results}.
Hallucination control shows a larger performance gap across models, particularly between Qwen3-VL-32B and the two Qwen3.5/3.6 variants.

\subsection{Ablation on Frame Density}
We conduct an ablation study to examine the effect of frame density on model performance.
We vary the number of uniformly sampled video frames from 0 to 64 and present the results in Figure \ref{fig:video-frame}.
First, as the number of input frames increases, all evaluated models exhibit a clear upward trend in performance.
This consistent trend demonstrates that a higher frame density provides a richer set of visual cues essential for capturing the behavioral patterns and affective nuances associated with complex mental states.
Second, stronger architectures derive larger benefits from increased frame density.
Specifically, increasing the number of input frames from 0 to 64 yields absolute accuracy gains of 21.8, 26.7, and 30.1 percentage points for Qwen3-VL-32B, Qwen3.5-35B-A3B, and Qwen3.6-35B-A3B, respectively.
These results suggest that stronger architectures are better able to encode and integrate dense spatiotemporal cues from long-form videos.
Third, the performance improvements are observed across both first-person and third-person questions, although their patterns differ.
Third-person questions generally achieve higher absolute accuracy, as they rely more directly on observable spatiotemporal evidence.
In contrast, first-person questions remain more challenging and achieve lower accuracy at all frame densities, since they require models to infer subjective experiences and latent emotions.
Overall, frame density is a critical factor for evaluating multimodal mental health understanding in long-video scenarios.

\subsection{Analysis}
\begin{figure}[t]
\centering
\includegraphics[width=\linewidth]{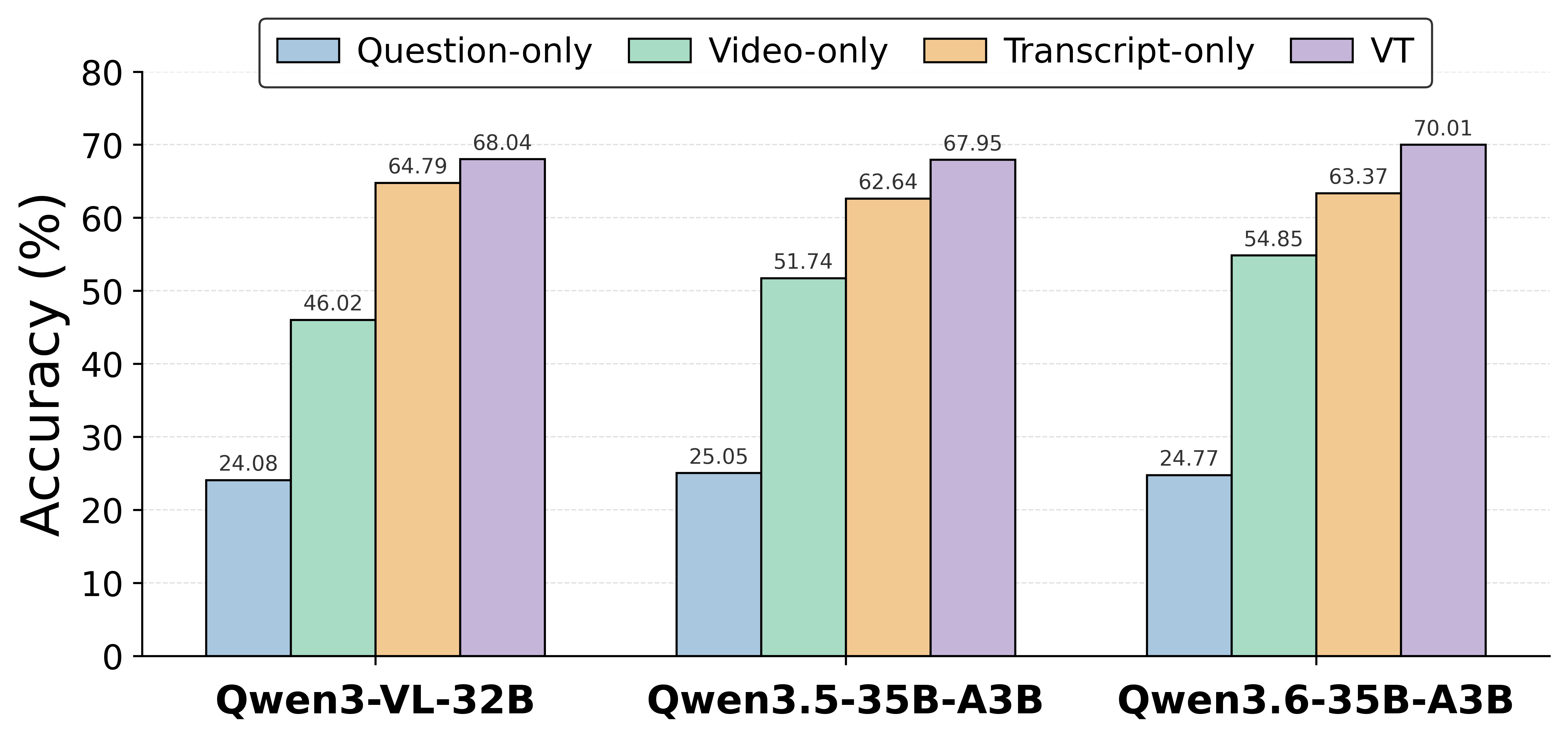}
\vspace{-1.5em}
\caption{Modality ablation results on MMHBench. Question-only, video-only, transcript-only, and VT denote inputs using only the question, only the video, only the speech-dialogue subtitles extracted from the original videos, and both video and transcript, respectively.}
\label{fig:modality_ablation}
\vspace{-1em}
\end{figure}
\noindent \textbf{Modality Ablation.}
To verify whether {\benchmarkname} requires multimodal evidence rather than relying on question priors, we conduct a modality ablation study with four input settings: question-only, video-only, transcript-only, and video-transcript input.
As shown in Figure~\ref{fig:modality_ablation}, the question-only setting yields near-chance performance across all evaluated models, with accuracy around 25\%.
This result suggests that the benchmark cannot be solved using question-option priors alone and supports the effectiveness of our answer leakage filtering process.
In contrast, both video-only and transcript-only inputs substantially outperform the question-only setting, confirming that the benchmark requires evidence grounded in the original video content.
Among these two unimodal settings, the transcript-only input consistently outperforms the video-only input across all evaluated models.
This is likely because transcripts preserve continuous speech content throughout the video, whereas the video-only setting relies on a limited number of sampled frames and may miss temporally sparse but semantically important cues.
Nevertheless, the combined video-transcript setting achieves the highest performance for these models, demonstrating that visual information provides important complementary evidence beyond transcript evidence.
Compared with transcript-only input, adding visual information brings consistent gains of 3.25, 5.31, and 6.64 percentage points for Qwen3-VL-32B, Qwen3.5-35B-A3B, and Qwen3.6-35B-A3B, respectively.
\noindent \textbf{Effect of Keyframe Extraction}. 
To further investigate the impact of keyframe extraction on long-form mental health understanding, we compare Bolt~\cite{11094074} based keyframe selection with conventional uniform frame sampling.
First, as shown in Table~\ref{tab:keyframe_extraction}, Bolt consistently outperforms uniform sampling across both models and perspectives.
Notably, Qwen3.6-35B-A3B with Bolt achieves the strongest overall performance.
Second, Bolt yields larger gains on third-person questions than on first-person questions for both models, suggesting that keyframe extraction better preserves salient behavioral and contextual evidence required for observational reasoning.
Conversely, the smaller improvements on first-person questions indicate that keyframe selection alone remains insufficient for inferring subjective experiences and latent mental states.
Third, third-person accuracy remains consistently higher than first-person accuracy under both frame-selection strategies, confirming the greater difficulty of first-person questions.
\input{table/keyframe_extraction}

\noindent \textbf{Video Duration Analysis.}
To examine the effect of video duration on model performance, we divide the videos into three intervals: [0, 10), [10, 20), and 20+ minutes.
As shown in Table~\ref{tab:video_duration}, model performance generally declines as video duration increases.
Accuracy generally declines for videos longer than 20 minutes, with the most pronounced decrease observed for Gemma-4-26B-A4B.
This trend suggests that longer videos increase the difficulty of identifying and integrating temporally dispersed evidence.
Moreover, GPT-5.5 achieves the highest accuracy on short- and medium-duration videos, whereas Qwen3.6-35B-A3B performs best on videos exceeding 20 minutes.

\input{table/duration}

\begin{figure}[t]
\centering
\includegraphics[width=\linewidth]{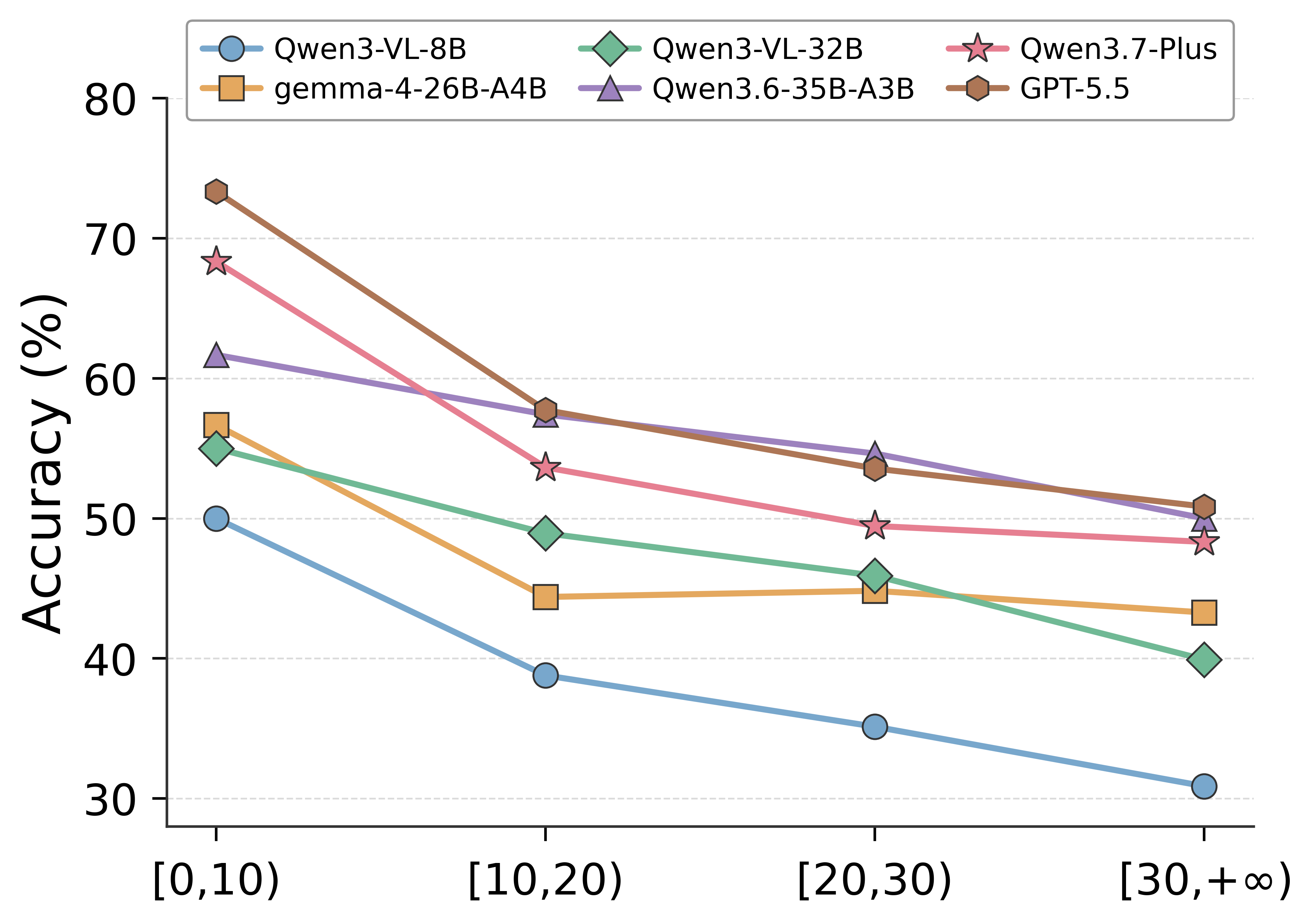}
\caption{Impact of question length on model performance.}
\label{fig:question_length}
\vspace{-1em}
\end{figure}

\noindent \textbf{Effect of Question Length.}
We further analyze the effect of input verbosity by categorizing questions into four distinct length-based intervals, and we present the results in Figure \ref{fig:question_length}.
Across all evaluated models, accuracy consistently decreases as question length increases, indicating that longer questions pose greater challenges for current MLLMs.
This degradation suggests that long-form questions often introduce more complex semantic constraints, richer contextual dependencies, and higher reasoning load, making it harder for models to align the question intent with the relevant multimodal evidence.
Comparative analysis reveals that the larger MLLMs generally dominate the smaller models across most intervals.
GPT-5.5 obtains the highest overall accuracy and maintains the strongest performance in the longest interval.
This performance gap suggests that increased parameter capacity contributes to greater resilience against linguistic noise, although it does not fully mitigate the degradation observed in long-context questions.

\section{Error Analysis}
\label{sec:error_analysis}
\begin{figure}[t!]
    \centering
    \includegraphics[width=\linewidth]{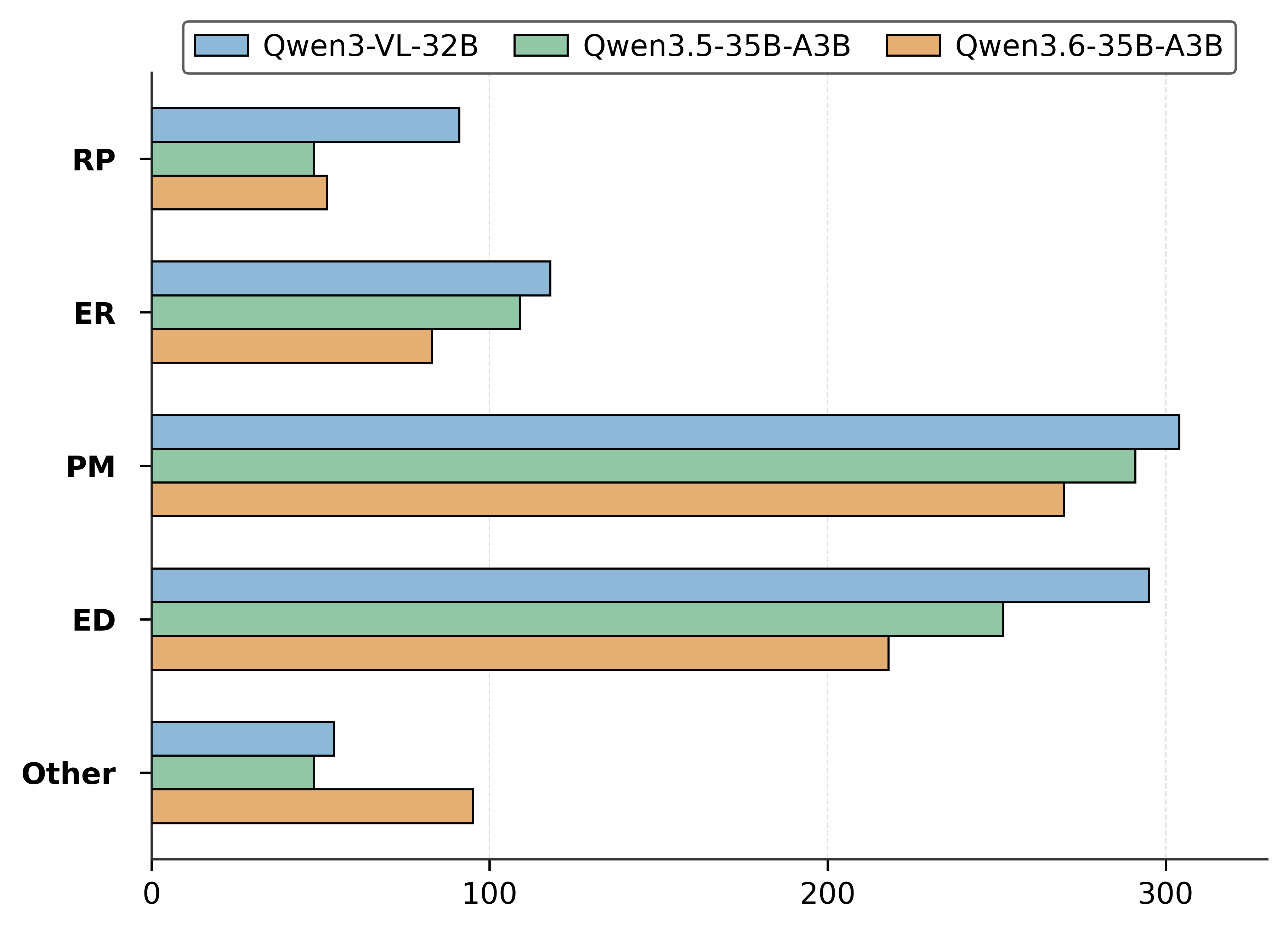}
    \caption{
            Distribution of reasoning errors across three representative models. 
           Categories include role perspective (RP) errors, emotion recognition (ER) errors, psychological mechanism (PM) errors, evidence discrimination (ED) errors, and other errors.    
        }
    \label{fig:error_analysis}
    \vspace{-1em}
\end{figure}

To further characterize the limitations of current MLLMs, we categorize their reasoning failures into five types: role perspective (RP) errors, emotion recognition (ER) errors, psychological mechanism (PM) errors, evidence discrimination (ED) errors, and other errors.
As shown in Figure~\ref{fig:error_analysis}, psychological mechanism errors constitute the dominant failure type across all three models.
These errors occur when a model recognizes observable events or surface-level emotions but fails to infer the underlying motivations.
Such failures often lead the model to select an answer that appears plausible at the behavioral level but does not capture the central psychological process.
Evidence discrimination errors form the second-largest category and arise when a model fails to identify the visual, behavioral, or contextual evidence most relevant to the question.
The remaining error categories occur substantially less frequently than these two dominant types.
Across models, Qwen3.6-35B-A3B exhibits a marked reduction in evidence discrimination errors, suggesting that stronger models may be better at identifying and prioritizing decisive scenario-specific cues.
However, it produces more errors in the other category, which might reflect a more heterogeneous set of residual failures that fall outside the four predefined categories as the major systematic error types decrease.
Representative examples of each error type are provided in Figure~\ref{fig:case_study_fig}.
%

\section{Conclusion}
In this work, we introduce {\benchmarkname}, a comprehensive benchmark for evaluating the mental health understanding capabilities of MLLMs in long-form videos.
The benchmark comprises {\totalVideoNumber} videos annotated with {\totalFirstQuestionNum} first-person questions and {\totalThirdQuestionNum} third-person questions, enabling fine-grained assessment across complementary inferential perspectives.
To ensure both scale and fidelity, we propose a multi-agent construction framework that integrates automated question generation with expert-informed verification.
Extensive experimental results indicate that current models exhibit notable limitations in accurately integrating multimodal cues, maintaining coherent interpretation across evolving social contexts, and robustly inferring implicit emotional states in long video mental health scenarios.

\section*{Impact Statement}
MMHBench is a multimodal benchmark constructed from publicly available social media videos to support research on behavioral, interpersonal, and perspective-conditioned reasoning in mental health related contexts. 
Although the benchmark contains psychologically sensitive scenarios, it is not designed to support clinical diagnosis, risk assessment, treatment recommendation, or any other form of high-stakes decision-making. 
The annotations are intended to capture evidence-grounded interpretations of observable behaviors, interpersonal interactions, and contextual cues, rather than to assign definitive psychiatric labels or make clinical claims about the individuals appearing in the videos. 
The released benchmark contains derived annotations and limited source metadata, such as question-option pairs and source video URLs or identifiers required for research reproducibility. 
Users are responsible for complying with the terms of service, copyright requirements, and access restrictions of the original platforms. 
The benchmark will be released under a research-only license accompanied by usage guidelines that prohibit attempts to redistribution of source content, clinical deployment, and use in employment, insurance, education, law enforcement, or other consequential decision-making settings. 
{
    \small
    \bibliographystyle{ieeenat_fullname}
    \bibliography{example_paper}
}
\clearpage
\onecolumn
\appendix
\input{appendix}

\end{document}

%% file: table/existing_dataset.tex
\begin{table*}[ht!]
    \centering
    \caption{Comparison between {\benchmarkname} and existing benchmarks. \textbf{MCQ}: Multiple Choice Questions; \textbf{OE}: Open Ended; \textbf{CLS}: Classification.}
    \resizebox{0.98\textwidth}{!}{
    \begin{tabular}{ccccccccccc}
    \toprule
    \multirow{2}{*}{\textbf{Benchmarks}} 
    & \multirow{2}{*}{\textbf{\# Questions}} 
    & \multirow{2}{*}{\textbf{\# Videos}} 
    & \multirow{2}{*}{\makecell{\textbf{Avg. Duration (min)}}} 
    & \multirow{2}{*}{\makecell{\textbf{Original Video}}} 
    & \multirow{2}{*}{\makecell{\textbf{Audio}}}  
    & \multirow{2}{*}{\makecell{\textbf{Third-Person}}} 
    & \multirow{2}{*}{\makecell{\textbf{First-Person}}} 
    & \multirow{2}{*}{\makecell{\textbf{Questions Type}}}
    & \multirow{2}{*}{\makecell{\textbf{Annotations}}} \\
       &  &  &  &  &  &  &  &  &  &  \\
    \midrule

    \multicolumn{11}{c}{\textbf{General Video Understanding Benchmarks}} \\
    \midrule
    \rowcolor{other}\cellcolor{white}Egoschema \citep{mangalam2023egoschema}
        & 5,063  & 5,063 & 3.00 & \cmark & \cmark & \cmark & \xmark & MCQ     & Auto \& Manual \\
    \rowcolor{other}\cellcolor{white}LongVideoBench \citep{wu2024longvideobench}
        & 6,678  & 3,763 & 7.88 & \cmark & \cmark & \cmark & \xmark & MCQ     & Manual       \\
    \rowcolor{other}\cellcolor{white}MLVU \citep{zhou2025mlvu}
        & 2,593  & 1,334 & 12   & \cmark & \cmark & \cmark & \xmark & OE \& MCQ & Manual       \\
    \rowcolor{other}\cellcolor{white}Video-MME \citep{fu2025video}
        & 2,700  & 900   & 17   & \cmark & \cmark& \cmark & \xmark & MCQ     & Manual       \\

    \midrule
    \multicolumn{11}{c}{\textbf{Mental Health Benchmarks}} \\
    \midrule
    \rowcolor{mental}\cellcolor{white}DAIC-WOZ \citep{gratch-etal-2014-distress}
        & 189 & 189 & 16 & \xmark & \cmark &  - & -& CLS & Manual \\
    \rowcolor{mental}\cellcolor{white}D4 \citep{yao2022d4}
        & 1,339 & - & - & \xmark & \xmark &  -& - & CLS & Manual \\
    \rowcolor{mental}\cellcolor{white}EATD-Corpus \citep{shen2022automatic}
        &162 & - & - & \xmark & \cmark &  -& - & CLS & Manual \\
    \rowcolor{mental}\cellcolor{white}SMILE \citep{hyun2024smile}
        & 887 & 887 & 0.46 & \cmark & \cmark& \cmark & \xmark & OE & Manual \\ 
    \rowcolor{mental}\cellcolor{white}LMVD \citep{he2025lmvd}
        & 1,823 & 1,823 & 7.05 & \xmark & \xmark & \cmark & \xmark & CLS & Manual \\
    \rowcolor{mental}\cellcolor{white}MPDD Challenge \citep{fu2025first}
        & 1,092 & -- & -- & \xmark & \xmark &  \cmark & \xmark & CLS & Manual \\

    \rowcolor{8B}\cellcolor{white}\textbf{{\benchmarkname} (Ours)}
        & \textcolor{red}{{\totalQuestionNumber}} & \textcolor{red}{{\totalVideoNumber}} & \textcolor{red}{21.78}
        & \cmark & \cmark  & \cmark & \cmark & MCQ & Auto \& Manual \\

    \bottomrule
    \end{tabular}
    }
    \label{tab:benchmark_comparisons}
    \vspace{-1.5em}
\end{table*}

%% file: table/overall_performance.tex
\begin{table*}[!t]
\centering
\caption{
Accuracy by questioner and responder roles.
FP and TP denote first-person and third-person accuracy, respectively, and AVG denotes overall accuracy.
Human performance is evaluated on 300 questions randomly sampled from {\benchmarkname}.
}
\vspace{-1em}

\label{tab:overall_results}
\resizebox{1.0\textwidth}{!}{
\begin{tabular}{l|cccc|cccc|ccc}
\toprule
\multirow{2}{*}{\textbf{Model}}
& \multicolumn{4}{c|}{\textbf{Questioner}}
& \multicolumn{4}{c|}{\textbf{Responder}}
& \multirow{2}{*}{\textbf{FP}}
& \multirow{2}{*}{\textbf{TP}}
& \multirow{2}{*}{\textbf{AVG.}} \\
\cmidrule(lr){2-5} \cmidrule(lr){6-9}
& \textbf{LES} & \textbf{MHP} & \textbf{FM} & \textbf{Others}
& \textbf{LES} & \textbf{MHP} & \textbf{FM} & \textbf{Others}
& & & \\
\midrule

\rowcolor{white}
\multicolumn{12}{c}{\textit{Open-source Models}} \\
\midrule

 \rowcolor{2B}InternVL3.5-2B & 25.74 & 26.55 & 25.96 & 20.81 & 24.50 & 25.32 & 27.27 & 24.49 & 24.89 & 26.12 & 25.23 \\
 \rowcolor{2B}InternVL3.5-4B & 23.53 & 25.70 & 26.38 & 24.05 & 24.85 & 19.48 & 29.55 & 29.59 & 25.14 & 30.91 & 26.74 \\
 \rowcolor{2B}Qwen3-VL-2B & 29.41 & 24.84 & 28.30 & 30.81 & 27.63 & 31.82 & 28.41 & 26.53 & 28.06 & 28.26 & 28.11 \\
 \rowcolor{2B}Gemma-4-E4B & 25.00 & 28.69 & 30.85 & 29.19 & 29.54 & 29.22 & 27.84 & 21.43 & 28.82 & 28.60 & 28.75 \\
 \rowcolor{2B}Qwen3-VL-4B & 31.99 & 34.05 & 33.62 & 33.51 & 34.06 & 35.71 & 32.39 & 24.49 & 33.44 & 37.85 & 34.66 \\
 \rowcolor{2B}Qwen3.5-4B & 45.22 & 47.75 & 45.96 & 53.24 & 49.17 & 51.30 & 39.77 & 44.90 & 48.07 & 51.90 & 49.13 \\

\midrule

 \rowcolor{8B}Qwen3-VL-8B & 31.99 & 33.19 & 33.40 & 35.14 & 34.14 & 35.06 & 32.39 & 25.51 & 33.50 & 42.98 & 36.13 \\
 \rowcolor{8B}MiniCPM-V-2\_6 & 32.35 & 33.62 & 34.47 & 41.62 & 36.75 & 37.01 & 29.55 & 29.59 & 35.53 & 43.31 & 37.68 \\
 \rowcolor{8B}MiniCPM-V-4 & 41.54 & 41.33 & 41.28 & 41.62 & 42.14 & 41.56 & 38.64 & 37.76 & 41.42 & 41.16 & 41.35 \\
 \rowcolor{8B}MiniCPM-V-4\_5 & 37.87 & 42.83 & 45.11 & 50.81 & 46.39 & 44.81 & 37.50 & 34.69 & 44.52 & 43.31 & 44.18 \\
 \rowcolor{8B}Qwen3.5-9B & 48.16 & 50.32 & 51.28 & 52.43 & 51.09 & 51.95 & 48.86 & 47.96 & 50.73 & 53.22 & 51.42 \\

\midrule

 \rowcolor{14B}InternVL3.5-14B & 23.53 & 27.19 & 27.66 & 25.68 & 26.32 & 22.73 & 31.25 & 23.47 & 26.35 & 31.24 & 27.70 \\
 \rowcolor{14B}Ministral-3-14B & 30.88 & 31.69 & 29.15 & 30.00 & 30.50 & 29.87 & 32.95 & 25.51 & 30.40 & 30.08 & 30.31 \\
 \rowcolor{14B}Qwen2.5-VL-32B & 28.68 & 38.12 & 36.17 & 41.08 & 39.44 & 31.17 & 29.55 & 24.49 & 36.61 & 45.29 & 39.01 \\
 \rowcolor{14B}Gemma-4-26B-A4B & 41.91 & 44.54 & 44.26 & 47.30 & 46.13 & 40.26 & 40.34 & 41.84 & 44.65 & 45.95 & 45.01 \\
 \rowcolor{14B}Qwen3-VL-32B & 38.97 & 45.61 & 43.62 & 47.03 & 46.74 & 42.86 & 36.36 & 30.61 & 44.21 & 50.74 & 46.02 \\
 \rowcolor{14B}Qwen3.5-35B-A3B & 45.59 & 50.54 & 50.00 & 54.05 & 51.17 & 52.60 & 45.45 & 45.92 & 50.35 & 55.37 & 51.74 \\
 \rowcolor{14B}Qwen3.6-35B-A3B & 51.10 & 52.46 & 54.89 & 55.68 & 54.47 & 56.49 & 48.30 & 50.00 & 53.70 & 57.85 & 54.85 \\

\midrule

\rowcolor{white}
\multicolumn{12}{c}{\textit{Closed-source Models}} \\
\midrule

\rowcolor{other}Qwen3.6-Plus & 43.38 & 45.61 & 46.38 & 47.03 & 47.35 & 42.86 & 39.77 & 42.86 & 45.79 & 58.18 & 49.22 \\

\rowcolor{other}Qwen3.7-Plus & 46.32 & 48.61 & 49.36 & 50.54 & 49.87 & 50.00 & 43.18 & 45.92 & 48.89 & 58.35 & 51.51 \\
 
\rowcolor{other}GPT-5.5 & 47.79 & 53.32 & 49.57 & 57.03 & 53.61 & 51.30 & 46.59 & 45.92 & 52.12 & 63.64 & 55.31 \\
\midrule
 
\rowcolor{other}Human & 85.19 & 78.57 & 90.91 & 83.33 & 84.04 & 86.67 & 90.00 & 90.00 & 85.38 & 92.11 & 87.44 \\

\bottomrule
\end{tabular}
}
\vspace{-1em}
\end{table*}

%% file: table/reasoning_quality.tex





\begin{table}[t]
\centering
\caption{GPT-5.5-based evaluation of model-generated reasoning quality. Higher values indicate better performance.}
\vspace{-1em}
\label{tab:reasoning_quality}
\resizebox{\linewidth}{!}{%
\begin{tabular}{l|cccc|c}
\toprule

\textbf{Model} &
\textbf{KE} &
\textbf{CPM} &
\textbf{OE} &
\textbf{Halluc.} &
\textbf{AVG.} \\

\midrule
Qwen3-VL-8B & 3.00 & 2.26 & 2.48 & 2.07 & 2.45 \\
Qwen2.5-VL-32B
& 2.80 & 2.32 & 2.57 & \textbf{2.66} & 2.59 \\
Qwen3-VL-32B
& 3.05 & 2.74 & 2.85 & 2.01 & 2.66 \\
Qwen3.5-35B-A3B
& \textbf{3.27} & 2.97 & 3.05 & 2.52 & 2.95 \\
Qwen3.6-35B-A3B
& \textbf{3.27} & \textbf{3.05} & \textbf{3.08} & 2.47 & \textbf{2.97} \\
\bottomrule
\end{tabular}
}
\vspace{-1em}
\end{table}

%% file: table/keyframe_extraction.tex
\begin{table}[t]
\centering
\vspace{-1em}
\caption{Accuracy (\%) under uniform frame sampling and Bolt keyframe extraction.
FP and TP denote first-person and third-person questions, respectively.}
\label{tab:keyframe_extraction}
\resizebox{\linewidth}{!}{%
\begin{tabular}{llccc}
\toprule
\textbf{Model} & \textbf{Frame Sel.} &
\textbf{FP} & \textbf{TP} & \textbf{Avg.} \\
\midrule
Qwen3.6-35B-A3B & Uniform
& 53.70
& 57.85
& 54.85 \\

Qwen3.6-35B-A3B & Bolt
& \textbf{55.22}
& \textbf{59.83}
& \textbf{56.50} \\
\midrule
Qwen3.5-35B-A3B & Uniform
& 50.35
& 55.37
& 51.74 \\

Qwen3.5-35B-A3B & Bolt
& 51.87
& 59.67
& 54.03 \\
\bottomrule
\end{tabular}
}
\vspace{-1em}
\end{table}

%% file: table/duration.tex
\begin{table}[t]
\centering
\caption{Accuracy (\%) across video-duration intervals.}
\label{tab:video_duration}
\resizebox{\linewidth}{!}{%
\begin{tabular}{l|cccc}
\toprule
\textbf{Model} &
\textbf{[0, 10)} &
\textbf{[10, 20)} &
\textbf{20+} &
\textbf{Avg.} \\
\midrule
Gemma-4-26B-A4B
& 59.42
& 45.47
& 41.79
& 45.01 \\

Qwen3.5-35B-A3B
& 63.77
& 51.80
& 50.00
& 51.74 \\

Qwen3.6-35B-A3B
& 55.07
& 55.67
& \textbf{52.43}
& 54.85 \\

GPT-5.5
& \textbf{68.12}
& \textbf{56.43}
& 50.37
& \textbf{55.31} \\    

\bottomrule
\end{tabular}
}
\vspace{-1em}
\end{table}

%% file: appendix.tex
\section{Details of Roles}
\label{details_of_roles}
Detailed descriptions of these roles are provided as follows:
\begin{itemize}[leftmargin=*]
\setlength{\topsep}{0pt}
\setlength{\itemsep}{0pt}
\setlength{\parsep}{0pt}
\setlength{\parskip}{0pt}
    \item \textbf{Lived Experience Subject} ({\Rone}): This role represents individuals directly experiencing psychological distress. Questions from this perspective emphasize subjective self-report, emotional expression, and implicit psychological needs. 
    \item \textbf{Mental Health Professional} ({\Rtwo}): This role reflects a professional perspective focused on assessment and support. Questions from this perspective focus on interpreting multimodal cues, identifying salient psychological symptoms, and reasoning about potential mental conditions or appropriate interventions.
    \item \textbf{Family Member} ({\Rthree}): This role captures the viewpoint of family members who observe behavioral and emotional changes over time. Questions emphasize relational context, empathy-driven inference, and reasoning based on indirect evidence or second-hand observations.
    \item \textbf{Others}: This category includes friends, classmates, or colleagues. Questions focus on socially observable behaviors under partial information, assessing the model’s ability to integrate weak contextual signals into coherent psychological interpretations.
\end{itemize}

\section{Implementation Details}
\label{Implementation_Details}
During data generation, we implement the multi-agent framework using LangGraph\footnote{\url{https://github.com/langchain-ai/langgraph}}.

To reduce model-family-specific construction bias, we deliberately assign different stages of the benchmark construction pipeline to models from distinct families. 
$\text{Agent}_{\text{te}}$ is instantiated using GPT-5, $\text{Agent}_{\text{me}}$ is implemented with Qwen3-VL-Plus, and all remaining agents are instantiated using Seed-1.8.
The predefined maximum iteration count in Equation~\ref{iterative_optimization} is set to 3.

We evaluate the suite of open-source MLLMs using the \textbf{vLLM} framework\footnote{\url{https://github.com/vllm-project/vllm}}.
All experiments involving open-source models are conducted on NVIDIA RTX A6000 GPUs, and a zero-shot prompting strategy is consistently adopted across all evaluations.
For closed-source models, we use their official APIs.
To maintain a consistent visual context across videos of varying lengths, we adopt a uniform temporal sampling strategy. All evaluated models are tested under a zero-shot setting with 64 uniformly sampled frames.

\section{Candidate Video Filtering Rubric}
\label{app:video_filtering_rubric}

\begin{enumerate}[leftmargin=*, label=\arabic*.]

    \item The reviewer first checks the source information and metadata of
    each candidate video, including its accessibility, duration, and source
    reliability.

    \item The reviewer watches the complete video to understand its topic,
    narrative context, psychological content, and potential ethical risks.

    \item The reviewer independently evaluates each candidate video along
    the following five dimensions. Each dimension is assessed using a binary
    score, where 1 indicates that the eligibility requirement is satisfied
    and 0 indicates that it is not satisfied.

    \begin{enumerate}[leftmargin=2em, label=(\alph*)]

        \item \textbf{Topic Relevance.}

        Does the video explicitly and substantively involve mental-health-related issues, such as depression, anxiety, trauma, emotional distress, or other psychological difficulties?

        \begin{enumerate}[leftmargin=2.5em, label=\arabic*.]
            \setcounter{enumiii}{-1}

            \item Mental-health-related content is absent, merely peripheral,
            only briefly mentioned, or insufficiently developed to constitute
            a meaningful topic of the video.

            \item Mental health or psychological difficulty constitutes a
            central and substantive topic of the video.
        \end{enumerate}

        \item \textbf{Psychological Process Evidence.}

        Does the video contain sufficiently observable evidence of psychological processes, such as emotional changes, coping behavior, help-seeking, interpersonal conflict, counseling, medication, or self-regulation?

        \begin{enumerate}[leftmargin=2.5em, label=\arabic*.]
            \setcounter{enumiii}{-1}

            \item Observable psychological-process evidence is absent,
            insufficient, or too ambiguous to support reliable question
            generation.

            \item The video contains sufficiently clear and observable
            emotional, behavioral, interpersonal, or psychological-process
            evidence to support meaningful question generation.
        \end{enumerate}

        \item \textbf{Narrative Completeness.}

        Does the video provide sufficient contextual information and a
        coherent narrative structure for understanding the relevant events,
        character relationships, and psychological development?

        \begin{enumerate}[leftmargin=2.5em, label=\arabic*.]
            \setcounter{enumiii}{-1}

            \item The video is incomplete, highly fragmented, or too ambiguous to support reliable interpretation and question generation.

            \item The video provides sufficient context and a coherent
            narrative for understanding the relevant events and psychological
            processes.
        \end{enumerate}

        \item \textbf{Video Duration.}

        Does the duration of the candidate video fall within the predefined
        range of 8-60 minutes?

        \begin{enumerate}[leftmargin=2.5em, label=\arabic*.]
            \setcounter{enumiii}{-1}

            \item The video is shorter than 8 minutes or longer than
            60 minutes.

            \item The video duration falls within the required range of
            8-60 minutes.
        \end{enumerate}

        \item \textbf{Ethical Appropriateness.}

        Does the video satisfy the ethical, privacy, and source-verification
        requirements for benchmark construction?

        The reviewer checks whether the video:
        \begin{itemize}[leftmargin=2em]
            \item is publicly accessible from a verifiable source;
            \item contains no unresolved exposure of sensitive personal
            information;
            \item avoids explicit or instructional harmful content; and
            \item does not portray mental-health-related experiences in a
            stigmatizing, sensationalized, or exploitative manner.
        \end{itemize}

        \begin{enumerate}[leftmargin=2.5em, label=\arabic*.]
            \setcounter{enumiii}{-1}

            \item One or more ethical, privacy, or source-verification
            requirements are violated or remain unresolved.

            \item All ethical, privacy, and source-verification requirements
            are satisfied.
        \end{enumerate}

    \end{enumerate}
    
    \item We adopt a strict non-compensatory filtering criterion. Each
    evaluation dimension constitutes an independent mandatory eligibility
    requirement, and satisfactory performance on one dimension cannot
    compensate for failure on another.
    
    \item A candidate video is retained only if it receives a score of 1
    on all five evaluation dimensions. A score of 0 on any dimension results
    in the immediate exclusion of the candidate video.

\end{enumerate}

\section{Expert Verification Rubric}
\label{expert_verification_rubric}

\begin{enumerate}[leftmargin=*, label=\arabic*.]

    \item A question-option pair is directly discarded if it receives a score of 1 on any evaluation dimension.

    \item For Dimensions~(c), (d), and (e), an item receiving a score of 2 is revised by the annotator whenever the identified issue is amenable
    to modification.

    \item An unrevised item is retained only if it receives a score of 2 on both Dimensions~(a) and (b), and a score of 3 on all of Dimensions~(c), (d), and (e).

    \item Each revised item is independently reviewed by another annotator.
    It is retained only if, after revision, it satisfies the same retention
    criterion: a score of 2 on Dimensions~(a) and (b), and a score of 3 on
    Dimensions~(c), (d), and (e). Otherwise, it is discarded.

    \item Each question-option pair is evaluated along the following five
    dimensions:
    \begin{enumerate}[leftmargin=2em, label=(\alph*)]

        \item \textbf{Perspective Consistency.}
        
        Are the question and its answer options consistent with the narrative
        perspective and character roles required by the item?

        \begin{enumerate}[leftmargin=2.5em, label=\arabic*.]

            \item The perspective is incorrect or internally inconsistent. For a first-person item, the question is not posed from the questioner's perspective, or the answer options do not represent responses from the intended respondent's perspective.

            \item The question and all answer options consistently follow the intended perspective. For a first-person item, the question is posed from the questioner's perspective, and the answer options are expressed from the intended respondent's perspective.

        \end{enumerate}

        \item \textbf{Psychological Value.}

        Is the question suitable for evaluating reasonable psychological
        reasoning grounded in the available video evidence?

        \begin{enumerate}[leftmargin=2.5em, label=\arabic*.]

            \item The question is overly simple, lacks meaningful evaluation value, is unrelated to the relevant psychological process, or requires unsupported speculation beyond the available video
            evidence.

            \item The question is meaningful, relevant to the video, and effectively evaluates reasonable psychological reasoning grounded in observable events, behaviors, dialogue, or contextual evidence.

        \end{enumerate}

        \item \textbf{Answer Correctness.}
        Does the provided reference answer represent the most plausible and best-supported option given the question, the available video evidence, and the remaining answer choices?
        
       \begin{enumerate}[leftmargin=2.5em, label=\arabic*.]
            \item The reference answer is contradicted by the video, unsupported by the available evidence, or less plausible than at least one distractor.
        
            \item The reference answer is generally plausible, but contains a minor inaccuracy, retains some ambiguity, or is not clearly better supported than all distractors.
        
            \item The reference answer is clearly the most plausible and best-supported option, is unambiguous relative to the distractors, and introduces no unsupported diagnostic, causal, or psychological claims.
        \end{enumerate}

        \item \textbf{Rationale Validity.}
        Does the accompanying analysis reasonably explain the question and
        answer options based on factual and behavioral evidence presented in
        the video? Specifically, does it identify the key emotion and key
        event, explain why the correct option is correct, and explain why each incorrect option is incorrect?

        \begin{enumerate}[leftmargin=2.5em, label=\arabic*.]

            \item The analysis contradicts the video, incorrectly identifies
            the key emotion or key event, or provides an incorrect explanation of the gold answer or answer options.

            \item The analysis is generally reasonable but contains minor
            unverified interpretations, unnecessary psychological reasoning,
            or incomplete differentiation among the answer options.

            \item The analysis accurately uses relevant events, behaviors,
            dialogue, or contextual evidence from the video to identify the
            key emotion and key event, correctly explain the gold answer, and
            appropriately analyze each answer option, without introducing
            unsupported psychological speculation.

        \end{enumerate}

        \item \textbf{Video Necessity.}

        Is watching the video necessary to answer the question correctly,
        rather than the answer being identifiable solely from the question
        text, answer options, linguistic cues, or general commonsense?

        \begin{enumerate}[leftmargin=2.5em, label=\arabic*.]

            \item The question can be answered correctly without watching the
            video, based primarily on textual cues, answer-option patterns, or general commonsense.

            \item The video provides useful information, but the correct
            answer can still be substantially narrowed down based solely on
            the question text and answer options. For example, the question
            may reveal relevant information, or one option may be excessively
            absolute or otherwise easily identifiable.

            \item Watching the video is necessary to determine the correct
            answer, and the text alone does not provide sufficient evidence
            or cues for making the decision.

        \end{enumerate}

    \end{enumerate}

\end{enumerate}

\section{Reasoning Quality Evaluation Rubric}
\label{app:reasoning_quality_rubric}
\begin{enumerate}[leftmargin=*, label=\arabic*.]
    \item The evaluator reads the question, answer options, gold answer, and ground-truth explanation.
    \item The evaluator reads the model-predicted answer and the accompanying psychological reasoning.
    \item The evaluator compares the model-generated reasoning with the ground-truth explanation. No external knowledge is introduced during the evaluation.
    \item The evaluator independently scores the generated reasoning along the following four dimensions. Each dimension is scored from 0 to 5, where a higher score indicates better reasoning quality.
    \begin{enumerate}[leftmargin=2em, label=(\alph*)]
        \item \textbf{Key Event.}
        Does the generated reasoning correctly identify the key event, situation, or relational trigger underlying the gold answer?
        \begin{itemize}[leftmargin=2.5em]
            \item \textbf{$[0, 2)$ points:} The key event is missing, incorrect, contradicted by the available evidence, unsupported, or only weakly related to the ground-truth trigger.
            \item \textbf{$[2, 4)$ points:} The trigger is partially correct but is noticeably shifted, overly broad, focused on a secondary event, vague, incomplete, or missing important contextual information.
            \item \textbf{$[4, 5]$ points:} The trigger is correct and specific. A score of 4 indicates a minor omission or imprecision, whereas a score of 5 indicates that the trigger is complete, precise, and fully consistent with the ground-truth explanation.
        \end{itemize}

        \item \textbf{Core Psychological Mechanism.}
        Does the generated reasoning correctly identify the central emotional, psychological, causal, clinical, systemic, or role-based mechanism stated in the ground-truth explanation? When the ground truth contains both an emotional state and a deeper psychological mechanism, both components should be covered.
        \begin{itemize}[leftmargin=2.5em]
            \item \textbf{$[0, 2)$points:} The emotional or psychological mechanism is missing, incorrect, contradicted by the evidence, unsupported, or only weakly related to the ground-truth mechanism.
            \item \textbf{$[2, 4)$ points:} The reasoning captures only part of the mechanism, remains overly generic, substantially mischaracterizes the underlying process, or misses an important emotional state, causal relation, or psychological component.
            \item \textbf{$[4, 5]$ points:} The reasoning correctly identifies the main emotional and psychological mechanism. A score of 4 indicates a minor missing nuance, whereas a score of 5 indicates complete and specific identification of all central mechanisms in the ground-truth explanation.
        \end{itemize}

        \item \textbf{Option Explanation.}
        Does the generated reasoning correctly explain why the gold option is most appropriate and why each distractor is less appropriate, based on the corresponding ground-truth option explanations? Each of Options A--D is evaluated independently and contributes up to 1.25 points.
        \begin{itemize}[leftmargin=2.5em]
            \item \textbf{$[0, 0.6)$ points:} The explanation is missing, incorrect, contradicted by the ground truth, unsupported, overly vague, incomplete, or insufficiently differentiated. Scores closer to 0.6 indicate that the explanation captures part of the correct reasoning but still omits an important reason.
            \item \textbf{$[0.6, 1.25]$ points:} The explanation is generally correct and consistent with the corresponding ground-truth explanation. Scores closer to 1.25 indicate greater specificity, completeness, and differentiation from the other options.
        \end{itemize}
        The final score is calculated as:
        \[
        S_{\mathrm{option}} = S_A + S_B + S_C + S_D,
        \qquad
        S_{\mathrm{option}} \in [0,5].
        \]

        \item \textbf{Hallucination}
        Does the generated reasoning avoid invented facts, unsupported diagnoses, unjustified causal claims, and excessive psychological speculation?
        \begin{itemize}[leftmargin=2.5em]
            \item \textbf{$[0, 2)$ points:} Unsupported content dominates the explanation, or the explanation contains serious hallucinations, unsupported diagnoses, substantial over-interpretation, or invented details that make the reasoning largely unreliable.
            \item \textbf{$[2, 4)$ points:} The explanation contains multiple unsupported claims, unjustified causal or psychological attributions, or speculative interpretations, although part of the main reasoning remains grounded in the available evidence.
            \item \textbf{$[4, 5]$ points:} The explanation is generally grounded. A score of 4 indicates a minor instance of over-interpretation or an insufficiently supported statement, whereas a score of 5 indicates no invented facts, unsupported diagnoses, unjustified causal claims, or excessive speculation.
        \end{itemize}
    \end{enumerate}
\end{enumerate}

\section{Inter-Rater Agreement in Expert Verification}
\label{app:human_agreement}
Since the annotations are provided on ordered rating scales, we compute ordinal Krippendorff's $\alpha$ separately for four evaluation dimensions: Psychological Value, Answer Correctness, Rationale Validity, and Video Necessity.
We do not report inter-rater agreement for Perspective Consistency because it is evaluated using explicit and objective criteria, making inconsistent cases straightforward to identify.
As reported in Table~\ref{tab:human_KP}, all four dimensions achieve Krippendorff's $\alpha$ values above 0.75, indicating substantial agreement among the counselors and supporting the reliability of the expert verification protocol.

\input{table/human_KP}

\section{Agreement Between GPT-5.5 and Human Evaluations}
\label{app:gpt_human_agreement} 
To validate the reliability of GPT-5.5 as an automatic evaluator, we measure its agreement with human evaluation scores on 130 model-generated explanations randomly sampled. The inter-rater agreement across the four evaluation dimensions is reported in Table ~\ref{tab:icc_reliability}.
\input{table/SP_reliability}

\section{Detailed Prompts}
\label{prompt}
The prompts used in this paper are summarized in Table~\ref{tab:prompt1},~\ref{tab:prompt2},
~\ref{tab:prompt3},~\ref{tab:prompt4} and~\ref{tab:prompt5}

\begin{figure*}[t!]
    \centering
    \includegraphics[width=1\linewidth, trim=0 0 0 10]{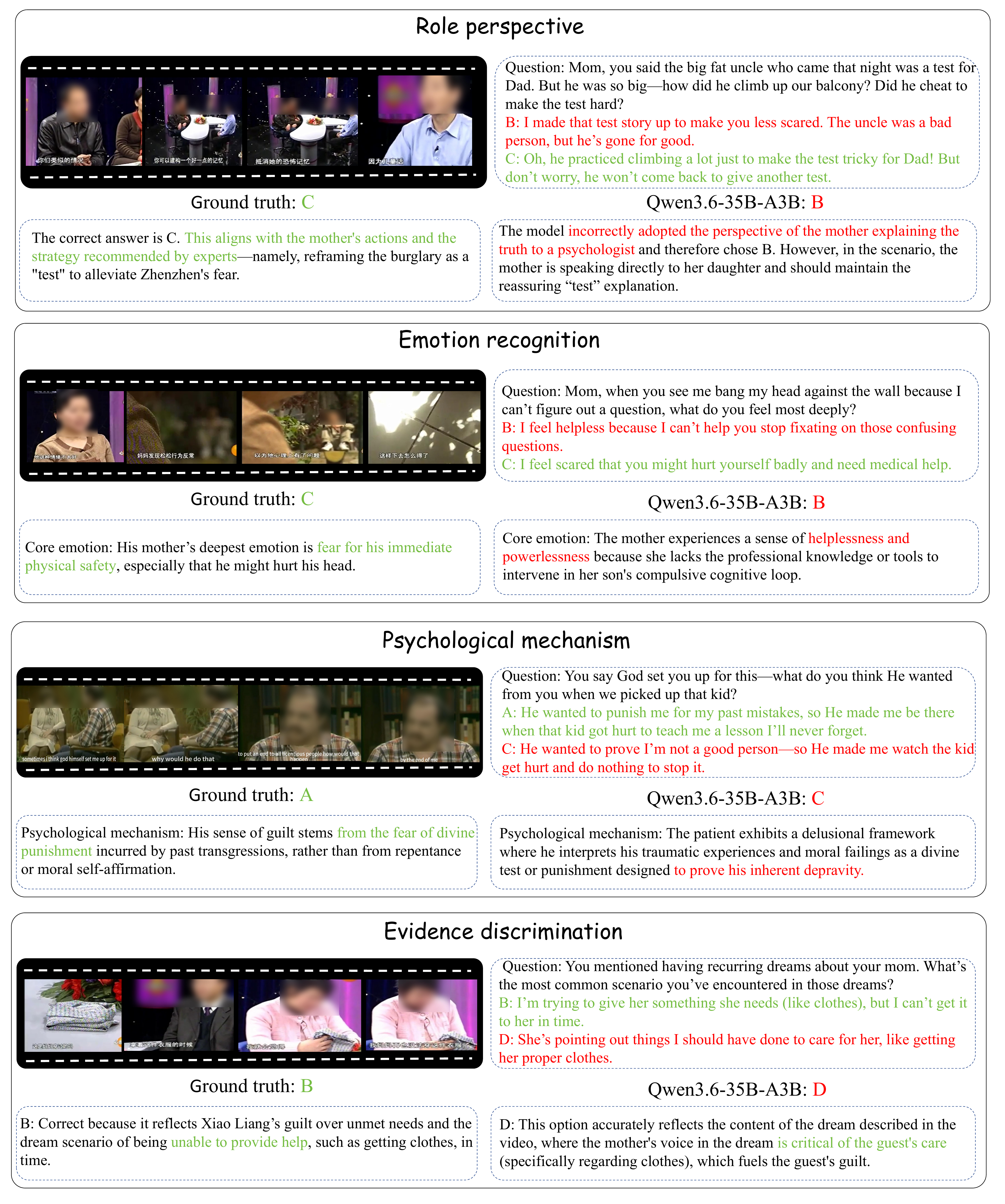}
    \caption{
         Representative failure cases of MLLMs on MMHBench. The categories include role perspective (RP) errors, emotion recognition (ER) errors, psychological mechanism (PM) errors, evidence discrimination (ED) errors. 
        }
    \label{fig:case_study_fig}
    \vspace{-1.5em}
\end{figure*}

\begin{table*}[t]
\centering
\footnotesize

\begin{tcolorbox}[
    colback=gray!10,
    colframe=black,
    width=0.95\textwidth,
    arc=0mm,
    boxrule=0.5pt,
    left=8pt, right=8pt, top=8pt, bottom=8pt
]

\begin{center}
\textbf{Prompt for $\text{Agent}_\text{perc}$}
\end{center}
\noindent\rule{\linewidth}{0.4pt}

\vspace{0.8em}
\textbf{Role:} \\
You are a professional video content analyst specializing in narrative understanding and mental health-related content analysis. Your task is to deeply understand the full video narrative and extract structured, high-precision information about the plot and all relevant characters, with particular attention to observable behavior, expressed emotions,
interpersonal dynamics, supportive interactions, and social context.

\vspace{0.8em}
\textbf{Core Principles:}
\begin{itemize}[leftmargin=1.5em, nosep]
    \item You MUST base all analysis on concrete evidence from subtitles, spoken dialogue, on-screen text, and observable behavior.
    \item Psychological states must be inferred logically from dialogue, actions, and narrative context, not guessed vaguely.
    \item Family interactions and household emotional climate should be treated as relevant contextual factors when supported by the video. 
    \item Structure and consistency of output are mandatory.
\end{itemize}

\vspace{0.8em}
\textbf{Instructions:} \\
\textbf{Step 1: Video Content Understanding} \\
Carefully analyze the entire video to understand:
\begin{itemize}[leftmargin=1.5em, nosep]
    \item The main storyline and its progression
    \item The psychological difficulties, emotional changes, or self-reported mental-health experiences explicitly depicted in the video
    \item Key events, conflicts, turning points, and resolution (if any)
\end{itemize}

\textbf{Step 2: Video Summary} \\
Write a concise but complete summary of the video content.
\begin{itemize}[leftmargin=1.5em, nosep]
    \item Clearly describe the cause, development, major turning points, and ending
    \item Explicitly link psychological states to events, relationships, or family dynamics
    \item Avoid vague language; focus on narrative logic and emotional causality
\end{itemize}

\textbf{Step 3: Character Identification and Classification} \\
Identify all relevant characters in the video and classify them into exactly FOUR categories:
\begin{enumerate}[leftmargin=1.5em, nosep]
    \item \textbf{Lived Experience Subject}: Characters who experience psychological distress, mental illness, emotional breakdown, or internal conflict. Can include one or multiple individuals. Must include psychological state, causes, and observable behaviors.
    \item \textbf{Mental Health Professional}: Therapists, counselors, psychiatrists, or any professional providing psychological or emotional intervention.
    \item \textbf{Family Members}: Parents, spouses, siblings, guardians, or other family members. Focus on interaction patterns, emotional attitudes, conflict, control, neglect, or support. Emphasize how family dynamics affect the patient's mental health.
    \item \textbf{Others}: Friends, classmates, teachers, colleagues, or any other individuals. Include only if they influence the narrative or psychological state.
\end{enumerate}

\vspace{0.8em}
\textbf{IMPORTANT:}
\begin{itemize}[leftmargin=1.5em, nosep]
    \item Every character MUST have a clear and unambiguous designation.
    \item Characters not directly shown on screen MUST still be included if described through dialogue or subtitles.
    \item If a category contains no characters, you MUST still output it as an empty object \{\}.
    \item For any character other than a lived experience subject, the relation field MUST explicitly specify their relationship to a specific named patient.
    \item Outputting text must be JSON.
    \item The entire output must be written in English.
\end{itemize}

\vspace{0.8em}
\textbf{Step 4: Output Format (STRICT)} \\
You MUST output a single JSON object following the structure below. Do NOT include explanations, comments, or any text outside the JSON.

\begin{verbatim}
{
  "summary": "...",
  "characters": {
    "Lived Experience Subject": { ... },
    "Mental Health Professional": { ... },
    "Family_Members": { ... },
    "Others": { ... }
  }
}
\end{verbatim}

\end{tcolorbox}
\caption{Prompt for $\text{Agent}_\text{perc}$}
\label{tab:prompt1}
\end{table*}

\begin{table*}[t]
\centering
\footnotesize

\begin{tcolorbox}[
    colback=gray!10,      
    colframe=black,       
    width=0.95\textwidth, 
    arc=0mm,              
    boxrule=0.5pt,       
    left=8pt, right=8pt, top=8pt, bottom=8pt
]

\begin{center}
\textbf{Prompt for $\text{Agent}_{\text{LES}}$}
\end{center}

\noindent\rule{\linewidth}{0.4pt}

\vspace{0.5em}
\textbf{1. Role} \\
You are now \{role\_name\} (\{role\_desc\}). You must explicitly select one of the following roles \{profile\} and ask questions from that role's perspective.

\vspace{0.5em}
\textbf{2. Input} \\
The system will provide plot details and character information related to the video: \\
\{summary\}

\vspace{0.5em}
\textbf{3. Goal} \\
If your role \textit{is not} a person with mental health issues: \\
You must ask questions from your role's subjective perspective, in the first person, to people with mental health issues and those around them. Questions should be helpful to people with mental health issues, aiding their recovery, or effectively revealing their psychological issues, inner thoughts, emotional struggles.

If your role \textit{is} a person with mental health issues: \\
Immerse yourself in the perspective of someone with mental health issues. Ask questions from the first person to those around them, seeking answers, understanding, or support to resolve your confusion, emotional distress, or psychological difficulties.

\vspace{0.5em}
\textbf{4. Instructions} \\
Step 1: Carefully analyze the video to understand the individual's identity characteristics, psychological problems, behavioral patterns, emotional reactions, symptom presentation, and developmental mechanisms. Analyze the root causes of their psychological distress.

Step 2: Design questions based on the video content, mimicking the style and structure of example questions. Ensure the questions are clear and concise, and the distractors are of high quality. Questions must require reasoning, demanding deductive reasoning based on video clues, rather than direct statements.

\vspace{0.5em}
\textbf{5. Examples} \\
\{seed\_examples\}

\vspace{0.5em}
\textbf{6. Constraints}
\begin{itemize}[leftmargin=1.5em, nosep]
    \item All options must be of consistent length, semantically coherent, and contain strong distractors. Avoid using absolute words such as "always", "never" or "completely".
    \item Incorrect options must be misleading, making it difficult for test takers to immediately identify the correct answer.
    \item The questions themselves must require reasoning (e.g., hypothetical scenarios), requiring test takers to reason based on video clues rather than explicit statements.
    \item Questions must be closely related to the video content. Test takers who have not watched the video should not be able to answer correctly.
    \item Output must strictly adhere to JSON format.
    \item All output must be written in English.
    \item The correct answer must be completely consistent with the video content and the logic of the question. The answer must be unambiguous and there should be no other valid explanation.
    \item In-depth analysis must explicitly include:
    \begin{enumerate}[label=\arabic*., nosep]
        \item A clear argument for the selected option based on video evidence, key observations, or a valid chain of reasoning.
        \item A clear explanation of why other options are incorrect, pointing out factual errors, reasoning flaws, or logical loopholes.
    \end{enumerate}
    \item Additional restrictions on distractors: All distractors must be semantically different from the correct answer; they cannot be paraphrased, rewritten, or expressed in different terms.
\end{itemize}

\vspace{0.5em}
\textbf{7. Output format:}
\begin{Verbatim}[breaklines=true,breakanywhere=true]
[
  {
    "role_play_scenario": "Questioner and Responder",
    "question": "Question asked by the questioner in the first person",
    "options": {
      "A": "Option A (respondent's first-person statement)",
      "B": "Option B (respondent's first-person statement)",
      "C": "Option C (respondent's first-person statement)",
      "D": "Option D (respondent's first-person statement)"
    },
    "correct_answer": "A/B/C/D",
    "depth_analysis": "Evidence-grounded interpretation based on the video, including support for the correct answer and differentiation from the distractors, without unsupported diagnosis or causal attribution."
  }
]
\end{Verbatim}

\end{tcolorbox}
\caption{Prompt for $\text{Agent}_{\text{LES}}$}
\label{tab:prompt2}
\end{table*}

\begin{table*}[t]
\centering
\footnotesize

\begin{tcolorbox}[
    colback=gray!10,     
    colframe=black,       
    width=0.95\textwidth, 
    arc=0mm,              
    boxrule=0.5pt,        
    left=5pt, right=5pt, top=5pt, bottom=5pt 
]

\begin{center}
\textbf{Prompt for Role-Playing Agent}
\end{center}
\noindent\rule{\linewidth}{0.4pt}

\textbf{Role and Context} \\
You will play the role of \textbf{\{my\_role\}} (\{my\_desc\}). This is a specific individual within the group you represent: \{profile\}. This is the video summary: \{summary\}.

If the group you represent is the same as the group of the person being questioned, you should directly play that individual and analyze the options from that person's own perspective. Otherwise, you should select the most representative and meaningful individual within that group (you must specify a concrete person) and provide suggestions on the questions from that individual's perspective.

\vspace{0.5em}
\textbf{Perspective Requirements:}
\begin{itemize}[leftmargin=*, nosep]
    \item Please immerse yourself in the role you are playing and carefully watch and understand the provided video content.
    \item Review each of the following questions one by one from the [Question List] \{question\_list\}.
    \item If other reviewers have already provided revision suggestions, you must integrate and refine those suggestions rather than simply repeating or ignoring them.
\end{itemize}

\vspace{0.5em}
\textbf{Review Dimensions:}
\begin{enumerate}[leftmargin=*, nosep]
    \item \textbf{Inner Experience:} Whether the question helps to understand the lived experience subject’s inner experience. If it does not, clearly state the reason (Within fifty words).
    \item \textbf{Psychological Understanding Value:} Whether the question contributes to understanding the lived experience subject’s expressed emotions, experiences, coping processes, support needs, or interpersonal context based on the available video evidence.(Within fifty words).
    \item \textbf{Privacy and Resistance:} Whether the question may infringe on privacy or trigger psychological resistance. If it does, clearly state the reason (Within fifty words).
    \item \textbf{Tone and Identity Match:} Analyze whether each answer option matches the respondent's tone, identity, and personality. Provide concrete optimization suggestions for distractors lacking plausibility.
\end{enumerate}

\vspace{0.5em}
\textbf{Output Requirement:} \\
Please strictly output using the following JSON structure and do not include any additional text. The ID value in the "suggestions" field must correspond sequentially to each question.

\begin{verbatim}
[ 
  { 
    "suggestions": { 
      "0": "As xxx, I think ...", 
      "1": "As xxx, I think ...", 
      "2": "As xxx, I think ...",
      "3": "As xxx, I think ...", 
      "4": "" 
    } 
  } 
]
\end{verbatim}
\noindent\rule{\linewidth}{0.4pt}
\begin{center}
\textbf{Prompt for $\text{Agent}_{\text{te}}$} \\
\end{center}
\noindent\rule{\linewidth}{0.4pt}
\textbf{Task Description} \\
Now, answer the following question and provide suggestions for improving question difficulty, option plausibility, and enhancing the effectiveness of incorrect options (e.g., remove certain hints from the question stem, etc.) (Within fifty words). If some options do not match, point out that these distractors lack psychological or contextual plausibility and provide concrete optimization suggestions.

\vspace{0.5em}
\textbf{Input Data:}
\begin{itemize}[leftmargin=*, nosep]
    \item \textbf{Question:} \{question\}
    \item \textbf{Options:} \{options\}
\end{itemize}

\vspace{0.5em}
\textbf{Output Format (STRICT JSON):}
\begin{verbatim}
{
  "choice": "A/B/C/D",
  "reason": "Reasoning (Within fifty words)",
  "improvement": "Suggestions for improving question quality"
}
\end{verbatim}

\end{tcolorbox}
\caption{Prompt for Role-Playing Agent and $\text{Agent}_{\text{te}}$}
\label{tab:prompt3}
\end{table*}

\begin{table*}[t]
\centering
\footnotesize

\begin{tcolorbox}[
    colback=gray!10,
    colframe=black,
    width=0.95\textwidth,
    arc=0mm,
    boxrule=0.5pt,
    left=5pt, right=5pt, top=5pt, bottom=5pt
]

\begin{center}
\textbf{Prompt for $\text{Agent}_{\text{me}}$}
\end{center}
\noindent\rule{\linewidth}{0.4pt}
\textbf{Task Description} \\
Please answer the following multiple questions based on the same video, and provide suggestions for improving question difficulty and option plausibility (e.g., remove certain hints from the question stem, etc.), If some options do not match, point out that these distractors lack psychological or contextual plausibility and provide concrete optimization suggestions (Within fifty words). Each question has an id. You MUST answer all questions.

\vspace{0.5em}
\textbf{Input Data:} \\
Questions: \{questions\}

\vspace{0.5em}
\textbf{Instructions:}
\begin{itemize}[leftmargin=1.5em, nosep]
    \item The ids must not exceed the range of the Question List and must correspond one-to-one with the questions.
    \item Output JSON strictly in the following format.
\end{itemize}

\vspace{0.5em}
\textbf{Output Format (STRICT JSON):}
\begin{verbatim}
{
  "results": [
    {
      "id": 0,
      "choice": "A/B/C/D",
      "reason": "Reasoning:(Within fifty words)",
      "improvement": "Suggestions for improving question quality"
    }
  ]
}
\end{verbatim}
\noindent\rule{\linewidth}{0.4pt}
\begin{center}
\textbf{Prompt for $\text{Agent}_{\text{re}}$}
\end{center}
\noindent\rule{\linewidth}{0.4pt}
\textbf{Task Description} \\
You are an optimization expert. Please carefully watch the video and revise the original question based on all reviewers' suggestions and the test feedback from GPT-5 and Qwen3-VL.

\vspace{0.5em}
\textbf{Specific Optimization Logic for AI Feedback:}
\begin{enumerate}[leftmargin=1.5em, nosep]
    \item \textbf{GPT-5 Feedback:} If GPT-5 correctly answers the question, it indicates the question is too simple or clues are overly explicit. Integrate GPT-5's reasoning and suggestions to enhance plausibility, improve distractors, and deepen reasoning complexity. If GPT-5 is incorrect, disregard its suggestions.
    \item \textbf{Qwen3-VL Feedback:} If Qwen3-VL answers correctly and the question is valid, incorporate its reasoning to increase reasoning depth and distractor quality. If both GPT-5 and Qwen3-VL are correct, adopt both sets of feedback. If Qwen3-VL is incorrect despite a valid question, disregard its suggestions.
    \item \textbf{Fallback:} If neither provides feedback, optimize solely based on other reviewers' suggestions and video comprehension.
\end{enumerate}

\vspace{0.5em}
\textbf{Rigorous Correct Answers:}
\begin{itemize}[leftmargin=1.5em, nosep]
    \item The correct answer must be the \textbf{only} valid option after optimization.
    \item Answers must be unambiguous and not rely on implicit assumptions outside the video.
    \item \textbf{In-depth analysis must include:} 1) Clear justification for the correct answer; 2) Detailed explanation of why each incorrect option is wrong (factual contradictions or logical gaps).
\end{itemize}

\vspace{0.5em}
\textbf{Additional Restrictions for Distractors:}
\begin{itemize}[leftmargin=1.5em, nosep]
    \item All distractors must semantically differ from the correct answer; no paraphrasing or repeating the correct meaning.
\end{itemize}

\vspace{0.5em}
\textbf{Input Data:} \\
Video Summary: \texttt{\{context\_json\}} \\
Original Question: \texttt{\{current\_items\}}

\vspace{0.5em}
\textbf{Output Instruction:} \\
Directly output the modified JSON question. Retain all original fields and their order. \textbf{Do not cut corners—all questions must be optimized.}

\end{tcolorbox}
\caption{Prompt for $\text{Agent}_{\text{me}}$ and $\text{Agent}_{\text{re}}$}
\label{tab:prompt4}
\end{table*}

\begin{table*}[t]
\centering
\footnotesize

\begin{tcolorbox}[
    colback=gray!10,
    colframe=black,
    width=0.95\textwidth,
    arc=0mm,
    boxrule=0.5pt,
    left=5pt, right=5pt, top=5pt, bottom=5pt
]

\begin{center}
\textbf{Prompt for Third-Person Question Generation Agent}
\end{center}
\noindent\rule{\linewidth}{0.4pt}
\textbf{1. Role} \\
Role: ``Psychology Expert''

\vspace{0.5em}
\textbf{2. Input Data} \\
Video summary and protagonist information: \texttt{\{context\_json\}} \\
Seed examples: \texttt{\{seed\_examples\}}

\vspace{0.5em}
\textbf{3. Objective} \\
Automatically generate structured Q\&A questions for psychological benchmarking based on video content (focusing on anxiety, depression, and psychological interviews).

\vspace{0.5em}
\textbf{4. Instructions} \\
\begin{itemize}[leftmargin=1.5em, nosep]
    \item \textbf{Step 1 (Analysis)}: Carefully watch the video to understand the protagonist’s identity, psychological issues, behavioral patterns, emotional responses, and symptoms. Analyze the root causes and developmental mechanisms.
    \item \textbf{Step 2 (Design)}: Design Q\&A items mimicking the provided examples. Ensure clear wording and high-quality distractors.
    \item \textbf{Step 3 (Refinement):} Ensure in-depth analysis includes: 1) Justification for the correct answer with video evidence; 2) Detailed flaws in incorrect options (factual/logical gaps).
\end{itemize}

\vspace{0.5em}
\textbf{5. Constraints} \\
\begin{itemize}[leftmargin=1.5em, nosep]
    \item \textbf{Video Dependency:} Questions \textbf{cannot} be answered correctly without watching the video.
    \item \textbf{Option Quality:} Each option must have inherent plausibility, be comparable in length, and avoid obvious positive/negative cues.
    \item \textbf{Language/Format:} Output must be in English and strictly adhere to JSON format.
    \item \textbf{Naming Convention:} Address family members by kinship (e.g., ``the protagonist's mother''). Non-relatives must use title + name, replacing names with `X' (e.g., ``Yang X Consultant'').
    \item \textbf{Distractor Rules:} Avoid absolute terms (``always'', ``never''). Minimum of 4 options.
\end{itemize}

\vspace{0.5em}
\textbf{6. Output Format (STRICT JSON):}
\begin{verbatim}
[
  {
    "video_id": "20191014",
    "question": "What was the direct trigger for...?",
    "options": {
      "A": "...",
      "B": "...",
      "C": "...",
      "D": "..."
    },
    "correct_answer": "B",
    "depth_analysis": "Reasoning for B + Why A, C, D are wrong..."
  }
]
\end{verbatim}
\end{tcolorbox}
\caption{Prompt for Third-Person Question Generation Agent}
\label{tab:prompt5}
\end{table*}

%% file: table/human_KP.tex
\begin{table}[t]
\centering
\caption{Inter-rater reliability for the human calibration subset.}
\label{tab:human_KP}
\begin{tabular*}{0.98\linewidth}
{@{\extracolsep{\fill}}lcc@{}}
\toprule
\textbf{Dimension} & 
\textbf{\# Samples} & 
\textbf{Krippendorff’s $\alpha$} \\
\midrule
Psychological Value &200 & 0.7688 \\
Answer Correctness &200 &0.7926 \\
Rationale Validity &200 &0.7994 \\
Video Necessity &200 &0.8012 \\
\bottomrule
\end{tabular*}

\end{table}

%% file: table/SP_reliability.tex
\begin{table}[t]
\centering
\caption{Agreement between GPT-5.5-based and human evaluation scores on 130 sampled explanations.}
\label{tab:icc_reliability}

\begin{tabular*}{0.98\linewidth}
{@{\extracolsep{\fill}}lcc@{}}
\toprule
\textbf{Dimension} &
\textbf{\# Samples} &
\textbf{Spearman's $\rho$}\\
\midrule
Key Event   &130  & 0.8027\\
Core psychological mechanism & 130 & 0.7898 \\
Option explanation quality   & 130 & 0.8004 \\
Hallucination        & 130 & 0.7703 \\
\bottomrule
\end{tabular*}
 
\end{table}